\documentclass{article}

\usepackage{microtype}
\usepackage{graphicx}
\usepackage{booktabs} 

\usepackage{subfigure}
\usepackage{xcolor}

\usepackage{hyperref}

\usepackage[accepted]{icml2023}

\usepackage{amsmath}
\usepackage{amssymb}
\usepackage{mathtools}
\usepackage{amsthm}
\usepackage{bm}

\usepackage{enumitem}

\usepackage{array}

\usepackage{makecell}
\usepackage[capitalize,noabbrev]{cleveref}

\theoremstyle{plain}
\newtheorem{theorem}{Theorem}[section]

\theoremstyle{definition}
\newtheorem{definition}[theorem]{Definition}

\theoremstyle{remark}

\usepackage[textsize=tiny]{todonotes}

\icmltitlerunning{On the Forward Invariance of Neural ODEs}

\begin{document}

\twocolumn[
\icmltitle{On the Forward Invariance of Neural ODEs}



\icmlsetsymbol{equal}{*}

\begin{icmlauthorlist}
\icmlauthor{Wei Xiao}{yyy}
\icmlauthor{Tsun-Hsuan Wang}{yyy}
\icmlauthor{Ramin Hasani}{yyy}
\icmlauthor{Mathias Lechner}{yyy}
\\\icmlauthor{Yutong Ban}{yyy}
\icmlauthor{Chuang Gan}{comp}
\icmlauthor{Daniela Rus}{yyy}

\end{icmlauthorlist}

\icmlaffiliation{yyy}{Computer Science and Artificial Intelligence Lab, Massachusetts Institute of Technology, Cambridge, MA, USA. }
\icmlaffiliation{comp}{MIT-IBM Watson AI Lab. Videos and code are available on the website: \url{https://weixy21.github.io/invariance/}}

\icmlcorrespondingauthor{Wei Xiao}{weixy@mit.edu}

\icmlkeywords{Neural ODEs, Robustness, Specification Guarantees, Safety}

\vskip 0.3in
]



\printAffiliationsAndNotice{}  

\begin{abstract}
We propose a new method to ensure neural ordinary differential equations (ODEs) satisfy output specifications by using invariance set propagation. Our approach uses a class of control barrier functions to transform output specifications into constraints on the parameters and inputs of the learning system. This setup allows us to achieve output specification guarantees simply by changing the constrained parameters/inputs both during training and inference.  
Moreover, we demonstrate that our invariance set propagation through data-controlled neural ODEs not only maintains generalization performance but also creates an additional degree of robustness by enabling causal manipulation of the system's parameters/inputs. We test our method on a series of representation learning tasks, including modeling physical dynamics and convexity portraits, as well as safe collision avoidance for autonomous vehicles.
\end{abstract}

\section{Introduction}
\label{sec:intro}
Neural ODEs \cite{chen2018neural} are continuous deep learning models that enable a range of useful properties such as exploiting dynamical systems as an effective learning class \cite{haber2017stable,gu2021efficiently}, efficient time series modeling \cite{rubanova2019latent,lechner2022mixed}, and tractable generative modeling \cite{grathwohl2018ffjord,liebenwein2021sparse}. 

\begin{figure}[t]
\begin{center}
\centerline{\includegraphics[width=\columnwidth]{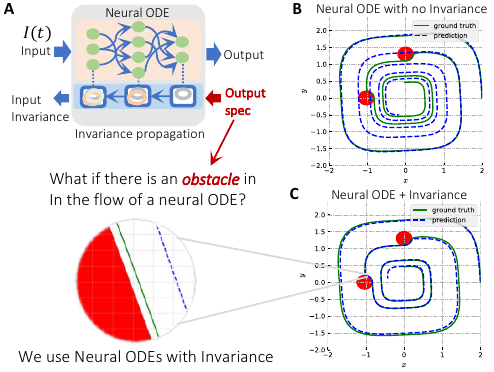}}
\caption{Invariance Propagation for neural ODEs. Output specifications can be guaranteed with invariance, including specification satisfaction between samplings, e.g., spiral curve regression with critical region avoidance. }
\label{fig:teaser}
\end{center}
 \vskip -0.3in
\end{figure}

Neural ODEs are typically trained via empirical risk minimization \cite{rumelhart1986learning, pontryagin2018mathematical}
endowed with proper regularization schemes \cite{massaroli2020dissecting} without much control over the behavior of the obtained network and over the ability to account for counterfactual inputs \cite{vorbach2021causal}. For example, a well-trained neural ODE instance that learned to chase a spiral dynamic (Fig. \ref{fig:teaser}\textbf{B}), would not be able to avoid an object on its flow, even if it has seen this type of \emph{output specification/constraint} during training. This shortcoming demands a fundamental fix to ensure the safe operation of these models specifically in safety-critical applications such as robust and trustworthy policy learning, safe robot control, and system verification \cite{lechner2020neural, kim2021stiff, hasani2022closed}.

In this paper, we set out to ensure neural ODEs satisfy output specifications. To this end, we introduce the concept of propagating invariance sets. An invariance set is a form of specification consisting of physical laws, mathematical expressions, safety constraints, and other prior knowledge of the structure of the learning task. We can ensure that neural ODEs are invariant to noise and affine transformations such as rotating, translating, or scaling an input, as well as to other uncertainties in training and inference.

To propagate invariance sets through neural ODEs we can use Lyapunov-based methods with forward invariance properties such as a class of control barrier functions (CBFs) \cite{Ames2017}, to formally guarantee that the output specifications are ensured. In order to account for non-linearity of the model, high-order CBFs \cite{Xiao2019}, a general form of CBFs, are required since high-relative degree constraints are introduced in such cases. CBFs perform this via migrating output specifications to the learning system's parameters or its inputs such that we can solve the constraints via forward calls to the learning system equipped with a quadratic program (QP). However, doing this requires a series of non-trivial novelties which we address in this paper. 1. CBFs are model-based Lyapunov methods, thus, they can only be used with data and systems with known dynamics. Here, we extend their formalism to work with unknown dynamics by the properties of neural ODEs. 2. CBFs are typically applied to systems with affine transformations. For neural ODEs with nonlinear activations, the propagation of invariance sets becomes a challenge. We fix this by incorporating a virtually linear space within the neural ODE to find simple parameter/input constraints.

Going back to Fig. \ref{fig:teaser}\textbf{C}, we observe that by applying our forward invariance propagation method, we can correct the model and force the system trajectories to stay away from the red obstacles while maintaining the path of the ground truth spiral curve. 



In summary, we make the following \textbf{new contributions}: 
\begin{itemize}
[align=right,itemindent=0em,labelsep=2pt,labelwidth=1em,leftmargin=*,itemsep=0em]
    \item We incorporate formal guarantees into neural ODEs via invariance set propagation.
    \item We use the class of Higher-order CBFs \cite{Xiao2021TAC2} to propagate the invariance set in neural ODEs while addressing their challenges such as handling unknown dynamics and the nonlinearity of the neural ODEs as well as connecting the order concept in HOCBFs to that of network depth in neural ODEs.
    \item We demonstrate the effectiveness of our method on a variety of learning tasks and output specifications, including the modeling of physical systems and the safety of neural controllers for autonomous vehicles. 
\end{itemize}

\section{Preliminaries}
In this section, we provide background on neural ODEs and forward invariance in control theory. 

\subsection{Neural ODEs}
\label{sec:bkg}

A neural ordinary differential equation (ODE) is defined in the form \citep{chen2018neural}:
\begin{equation}\label{eqn:NN}
\dot {\bm x}(t) = f_{\theta}(\bm x(t)),
\end{equation}
where $n\in \mathbb{N}$ is state dimension, $\bm x \in \mathbb{R}^n$ is the state and $\dot{\bm x}$ denotes the time derivative of $\bm x$, $f_{\theta}:\mathbb{R}^n \rightarrow\mathbb{R}^n$ is a neural network model parameterized by $\theta$. 
 The output of the neural ODE is the integral solution of (\ref{eqn:NN}). 
It can also take in external input, where the model is defined as:
\begin{equation}\label{eqn:NN_control}
\dot {\bm x}(t) = f_{\theta}^{'}(\bm x(t), \textbf{I}(t)),
\end{equation}
where $n_\textbf{I} \in\mathbb{N}$ is external input dimension, $\textbf{I}(t) \in\mathbb{R}^{n_\textbf{I}}$, $f_{\theta}^{'}:\mathbb{R}^n\times \mathbb{R}^{n_\textbf{I}} \rightarrow\mathbb{R}^n$ is a neural network model parameterized by $\theta$. For notation convenience, we write as,
\begin{equation}
\label{eqn:NN_decomp}
f_\theta = f_{\theta_{K,K+1}} \circ \cdots \circ f_{\theta_{1,2}}
\end{equation}
where $K$ is the number of layers, $f_{\theta_{k,k+1}}, k\in[1,K]$ is the forward process of the $k$'th layer, and we denote $\bm z_k=(f_{\theta_{k,k+1}} \circ \cdots \circ f_{\theta_{1,2}})(\cdot)\in \mathbb{R}^{n_k}$ the intermediate representation at the $k$'th layer and $\bm z_K=\dot{\bm x}$ is the output. $n_k\in\mathbb{N}$ denotes the number of neurons at layer $k$ and $n_K = n$.


\subsection{Forward Invariance in Control Theory}

Consider an affine control system of the form
\begin{equation}
\dot{\bm{x}}=f(\bm x)+g(\bm x)\bm u \label{eqn:affine}%
\end{equation}
where $\bm x\in\mathbb{R}^{n}$, $f:\mathbb{R}^{n}\rightarrow\mathbb{R}^{n}$
and $g:\mathbb{R}^{n}\rightarrow\mathbb{R}^{n\times q}$ are {locally}
Lipschitz, and $\bm u\in U\subset\mathbb{R}^{q}$, where $U$ denotes a control constraint set. 

\begin{definition}
    (\textbf{Set invariance}):
    \label{def:forwardinv} 
    A set $C\subset\mathbb{R}^{n}$ is forward invariant for system (\ref{eqn:affine}) if its solutions for some $\bm u\in U$ starting at any $\bm x(t_0) \in C$ satisfy $\bm x(t)\in C,$ $\forall t\geq t_0$.
\end{definition}

\begin{definition}
    \label{def:relative} 
    (\textbf{Relative degree}): 
    The relative degree of a differentiable function $b:\mathbb{R}^{n}\rightarrow\mathbb{R}$ (or constraint $b(\bm x)\geq 0$) with respect to the system (\ref{eqn:affine}) is the number of times $b(\bm x)$ needs to be differentiated along dynamics (\ref{eqn:affine}) until any component of $\bm u$ explicitly shows in the corresponding derivative.
\end{definition}



\begin{definition}
    \label{def:class_k} 
    (\textbf{Class $\mathcal{K}$ function}): 
    A Lipschitz continuous function $\alpha: [0,a) \rightarrow [0,\infty), a>0$ belongs to class $\mathcal{K}$ if it is strictly increasing and $\alpha(0)=0$.
\end{definition}

\begin{definition}
    \label{def:hobf} 
    (\textbf{High Order Barrier Function (HOBF)}):
    A function $b: \mathbb{R}^{n}\rightarrow\mathbb{R}$ of relative degree $m$ is a HOBF with a sequence of functions $\psi_i: \mathbb{R}^n \rightarrow \mathbb{R}$ such that $\psi_m(\bm x)\geq 0$,
    \begin{equation}
    \begin{aligned} \psi_i(\bm x) := \dot \psi_{i-1}(\bm x) + \alpha_i(\psi_{i-1}(\bm x)),\quad i\in\{1,\dots,m\}, \end{aligned} \label{eqn:functions}%
    \end{equation}
    where $\psi_{0}(\bm x):=b(\bm x)$ and $\alpha_{i}(\cdot)$ is a $(m-i)$'th order differentiable class $\mathcal{K}$ function. We define a sequence of sets,
    \begin{equation}
    \label{eqn:sets}\begin{aligned} C_i := \{\bm x \in \mathbb{R}^n: \psi_{i-1}(\bm x) \geq 0\}, \quad i\in\{1,\dots,m\}. \end{aligned}
    \end{equation}
\end{definition}

\begin{definition}
    \label{def:hocbf}
    (\textbf{High Order Control Barrier Function (HOCBF)} \cite{Xiao2021TAC2}):
    Let $\psi_i$ and $C_i$ be defined by \eqref{eqn:functions} and \eqref{eqn:sets}, respectively, for $i\in\{1,\dots,m\}$. A function $b:\mathbb{R}^n\rightarrow\mathbb{R}$ is a HOCBF of relative degree $m$ if there exists $(m-i)^{th}$ order differentiable class $\mathcal{K}$ functions $\alpha_{i},i\in\{1,\dots,m\}$ such that
    \begin{equation}
    \label{eqn:constraint}
    \begin{aligned} 
    \sup_{\bm u\in U}[L_f^{m}b(\bm x) + [L_gL_f^{m-1}b(\bm x)]\bm u \!+\! O(b(\bm x)) \\+ \alpha_m(\psi_{m-1}(\bm x))] \geq 0, 
    \end{aligned}
    \end{equation}
    for all $\bm x\in C_{1} \cap,\dots, \cap C_{m}$. $L_{f}$ and $L_{g}$ denote Lie derivatives w.r.t. $\bm x$ along $f$ and $g$, respectively, and $O(b(\bm x)) = \sum_{i = 1}^{m-1}L_f^i(\alpha_{m-i}\circ\psi_{m-i-1})(\bm x)$. The satisfaction of (\ref{eqn:constraint}) is equivalent to the satisfaction of $\psi_m(\bm x)\geq 0$ defined in (\ref{eqn:functions}).
\end{definition}

The HOCBF is a general form of the CBF \cite{Ames2017} (a HOCBF with $m = 1$ degenerates to a CBF), and it can be applied to arbitrary relative degree systems, such as the invariance propagation to nonlinear layers of a neural ODE in this work.

\begin{theorem}
    \label{thm:hocbf} 
    \cite{Xiao2021TAC2}):
    Given a HOCBF $b(\bm x)$ from Def. \ref{def:hocbf} with the sets $C_{1}, \dots, C_{m}$ defined by (\ref{eqn:sets}), if $\bm x(t_0) \in C_{1} \cap,\dots,\cap C_{m}$, then any Lipschitz continuous controller $\bm u(t)$ that satisfies the constraint in (\ref{eqn:constraint}), $\forall t\geq t_0$ renders $C_{1}\cap,\dots, \cap C_{m}$ forward invariant for system (\ref{eqn:affine}).
\end{theorem}


In this work, we map the forward invariance in control theory to \textit{forward invariance in neural ODEs}, where we tackle arbitrary dynamics defined by neural ODE $f_\theta$ (which can be nonlinear as opposed to an affine control system in the form of (\ref{eqn:affine})).





\section{Invariance Propagation}
\label{sec:ip}

In this section, we present the theoretical framework of \textit{Invariance Propagation (IP)} to guarantee forward invariance (in short, invariance) of a neural ODE. We first provide formalisms of the proposed method. Then, we describe invariance propagation to (i) linear layer (ii) nonlinear layer (iii) external input. 



\noindent\textbf{Output Specification.} A continuously differentiable function $h: \mathbb{R}^n \rightarrow \mathbb{R}$ constructs an output specification $h(\bm x) \geq 0$ for a neural ODE. Typical output specifications include system safety (e.g., collision avoidance in autonomous driving), physical laws (e.g., energy conservation), mathematical formulae (e.g., Cauchy Schwarz inequality), etc.


\begin{definition}
    (\textbf{Forward Invariance in Neural ODE}):
    \label{def:inv}
    The (forward) invariance of a neural ODE \eqref{eqn:NN} or \eqref{eqn:NN_control} with $f_\theta$ is defined w.r.t. its output specification $h(\bm x)\geq 0$ such that if $h(\bm x(t_0)) \geq 0$, then $h(\bm x(t))\geq 0, \forall t\geq t_0$, where $x(t)=\int_{t_0}^t f_\theta(\tau) d\tau$. Intuitively, this property guarantees the satisfaction of output specification is forwarded in the neural ODE across time.
\end{definition}



\begin{definition}
    (\textbf{Invariance Propagation} (IP)):
    Given an output specification $h(\bm x) \geq 0$ and a neural ODE $f_\theta$, invariance propagation describes a procedure to find a constraint $\Psi(\bm d)\geq 0$, where $\Psi:\mathbb{R}^{n_{\bm d}}\rightarrow \mathbb{R}$ and $\bm d$ ($n_d\in\mathbb{N}$ is its dimension) is either (i) a subset of parameters $\theta$, (ii) the external input $\textbf{I}$, or (iii) other auxiliary variables for the neural ODE, such that if $\Psi(\bm d)\geq 0$, forward invariance defined in  Def. \ref{def:inv} is satisfied. Intuitively, IP casts invariance w.r.t. output specification to pose constraints on non-output in ODEs.
\end{definition}


\subsection{Invariance Propagation to Linear Layers}
\label{sec:output}
We start with a simple case where invariance is propagated to linear layers of the neural ODE, which normally occurs at the output layer without nonlinear activation functions.


\noindent\textbf{Neural ODE Reformulation.} Without loss of generality, we follow \eqref{eqn:NN_decomp} and assume a linear output layer $f_{\theta_{K-1,K}}$,
\begin{equation}
    \label{eqn:NN_affine}
    \dot {\bm x} = \sum_{i=1}^{n_2}\theta_{K-1,K}^i {\bm z}_{K-1}^i = \theta_{K-1,K}^{\mathcal{P}}{\bm z}_{K-1}^{\mathcal{P}} + \theta_{K-1,K}^{\mathcal{N}}{\bm z}_{K-1}^{\mathcal{N}}
\end{equation}
where ${\bm z}_{K-1}=(f_{\theta_{K-1,K-2}}\circ \cdots \circ f_{\theta_{1,2}})(\bm x)$, $\theta_{K-1,K}^i$ is the $i$'th column of $\theta_{K-1,K}\in \mathbb{R}^{n\times n_{K-1}}$, ${\bm z}_{K-1}^i$ is the $i$'th entry of ${\bm z}_{K-1}\in\mathbb{R}^{n_{K-1}}$, $\mathcal{P}$ and $\mathcal{N}$ describe sets of columns that are updatable parameters (that the invariance is propagated to) and constants, respectively. We drop the bias term for cleaner notation.

 \textbf{Propagation to Linear Layers.} Our goal is to propagate the invariance to a subset of parameters. We treat $\theta_{K-1,K}^{\mathcal{P}}$ as a variable while taking other parameters $\theta_{K-1,K}^{\mathcal{N}}$ as constants. Given an arbitrary output specification $h(\bm x)\geq 0$, we can define a $\psi_1$ function in the form:
\begin{equation} \label{eqn:function1}
    \psi_1(\bm x, \theta_{K-1,K}^{\mathcal{P}}):= \frac{dh(\bm x)}{d\bm x} f_{\theta}(\bm x) + \alpha_1(h(\bm x)),
\end{equation}
where $\alpha_1(\cdot)$ is a class $\mathcal{K}$ function. Note that $\theta_{K-1,K}^{\mathcal{P}}$ is implicitly defined in $f_{\theta}$.
Combining \eqref{eqn:function1} with \eqref{eqn:NN_affine}, the following theorem shows the invariance of the neural ODE (\ref{eqn:NN}):
\begin{theorem}\label{thm:inv}
Given a neural ODE as in (\ref{eqn:NN_affine}) and an output specification $h(\bm x) \geq 0$, if there exist a class $\mathcal{K}$ function $\alpha_1$ and $\theta_{K-1,K}^{\mathcal{P}}$ such that with $\psi_1$ as in \eqref{eqn:function1}, 
\begin{equation} \label{eqn:inv}
\begin{split}
 \Psi(\theta_{K-1,K}^{\mathcal{P}}|\bm x) = \psi_1(\bm x, \theta_{K-1,K}^{\mathcal{P}}) \geq 0,
\end{split}
\end{equation}
for all $\bm x$ such that $h(\bm x)\geq 0$, 
where $\Psi(\theta_{K-1,K}^{\mathcal{P}}|\bm x) = \frac{dh}{d \bm x}\theta_{K-1,K}^{\mathcal{P}}{\bm z}_{K-1}^{\mathcal{P}} + \frac{dh}{d \bm x}\theta_{K-1,K}^{\mathcal{N}}{\bm z}_{K-1}^{\mathcal{N}} + \alpha_1(h(\bm x))$,
then the neural ODE is forward invariant.
\end{theorem}
The proof and existence of $\alpha_1$ are shown in Appendix \ref{asec:proof1}.

\textbf{Brief Summary.} Thm. \ref{thm:inv} provides a condition on the parameter $\theta_{K-1,K}^{\mathcal{P}}$ that implies the invariance of the neural ODE. In other words, by modifying the parameter $\theta_{K-1,K}^{\mathcal{P}}$ such that (\ref{eqn:inv}) is always satisfied, we can guarantee the invariance. The algorithm is shown in the next section. Moreover, since we only need to take the derivative of $h(\bm x)$ once, as shown in (\ref{eqn:inv}), this is analogous to a first-order HOCBF (i.e., $m=1$ in Def. \ref{def:hocbf}).

\textbf{IP to the Proper Parameters.} By using the Theorem \ref{thm:inv} , we can propagate the invariance to neural ODE parameters in linear layers.  However, note that we need to choose the parameters such that all the output of the neural ODE are able to be changed by modifying the target parameters. Otherwise, the output specification may fail to be guaranteed. 


\subsection{Invariance Propagation to Nonlinear Layers}
\label{sec:param}
In this section, we consider how we may efficiently propagate the invariance to the weight parameters of an arbitrary layer of the neural ODE (including the output layer with nonlinear activation functions). Theoretically, we can propagate the invariance to arbitrary layers using the existing HOCBF theory. However, the resulting invariance enforcement would be nonlinear programs (i.e., the HOCBF constraint (\ref{eqn:constraint}) will be nonlinear in $\bm u$), which are computationally hard and inefficient to solve. Moreover, the formulation above does not allow us to incorporate the IP in the training loop to address the conservativeness of the invariance as discussed next. Our method works for both (\ref{eqn:NN}) and (\ref{eqn:NN_control}), so we only consider (\ref{eqn:NN}) for simplicity.

\begin{figure}[t!]
\begin{center}
\centerline{\includegraphics[width=\columnwidth]{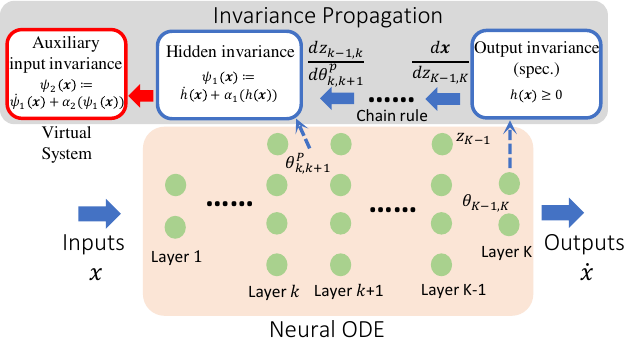}}
\caption{Invariance propagation to an arbitrary layer of the neural ODE with auxiliary virtual linear space. }
\label{fig:prop}
\end{center}
\end{figure}

\textbf{An Auxiliary Linear System.} Given a neural ODE \eqref{eqn:NN}, we want to propagate the invariance to the partial parameter (similar to \eqref{eqn:NN_affine}) at the $k$'th layer $\theta_{k, k+1}^{\mathcal{P}}\in\mathbb{R}^{  n_{k+1}^\mathcal{P}\times n_k }$ with $k\in\{1,\dots, K\}$ and $n_{k+1}^\mathcal{P}\leq n_{k+1}$. Then, we flatten the matrix parameter $\theta_{k, k+1}^\mathcal{P}$ to a vector form $\theta_{k}^\mathcal{P} \in\mathbb{R}^{d_{k}^\mathcal{P}}$ row-wise, where $d_{k}^\mathcal{P} = n_{k+1}^\mathcal{P} n_k$ is the dimension of the vector.
Instead of directly propagating the invariance to the parameter $\theta_{k}^\mathcal{P}$ and resulting nonlinear constraints, we propagate the invariance to an auxiliary linear system:
\begin{equation} \label{eqn:linear}
    \dot \theta_{k}^\mathcal{P} = A_{k}^\mathcal{P} \theta_{k}^\mathcal{P} + B_{k}^\mathcal{P} u_{k}^\mathcal{P}
\end{equation}
where $A_{k}^\mathcal{P}\in\mathbb{R}^{d_k^\mathcal{P}\times d_k^\mathcal{P}}, B_{k}^\mathcal{P}\in\mathbb{R}^{d_k^\mathcal{P} \times d_k^\mathcal{P}}$ are chosen such that the auxiliary system is controllable and $u_{k}^\mathcal{P}\in\mathbb{R}^{d_k^\mathcal{P}}$ is the auxiliary control input. The exact choice of $A_{k}^\mathcal{P}$ and $B_{k}^\mathcal{P}$ may slightly impact the performance, which is further discussed in Appendix \ref{asec:aux}. This specific formulation allows performing IP on $u_{k}^\mathcal{P}$ linearly (which will become clearer later on) as opposed to directly on $\theta_{k}^\mathcal{P}$, which is susceptible to nonlinearity. An overview is illustrated in Fig. \ref{fig:prop}.


\textbf{Propagation to Auxiliary System.} We first propagate the invariance to parameter $\theta_{k}^\mathcal{P}$ by defining a function $\psi_{1}$ similar to (\ref{eqn:function1}), which is illustrated by the blue boxes in Fig. \ref{fig:prop}. Then, we further propagate the invariance to the $u_k^p$ in system (\ref{eqn:linear}) by defining another function $\psi_2$ (the red box in Fig. \ref{fig:prop}):
\begin{equation} \label{eqn:function2}
\begin{aligned}
  \psi_1(\bm x, \theta_k^\mathcal{P}) &:=\frac{dh(\bm x)}{d\bm x}f_{\theta}(\bm x)  + \alpha_1(h(\bm x)),  \\ \psi_2(\bm x, u_k^\mathcal{P}) &:=\frac{\partial \psi_1}{\partial\bm x} f_{\theta}(\bm x) +  \frac{\partial\psi_1}{\partial\theta_k^\mathcal{P}} \dot \theta_k^\mathcal{P}+ \alpha_2(\psi_1(\bm x, \theta_k^\mathcal{P})),
  \end{aligned}
\end{equation}
where $\alpha_1(\cdot), \alpha_2(\cdot)$ are class $\mathcal{K}$ functions and $\psi_1,\psi_2$ are defined in the similar spirit to \eqref{eqn:functions}. Remark that different from \eqref{eqn:function1}, here, $\psi_1$ is nonlinear regarding $\theta_k^\mathcal{P}$ yet the newly introduced $\psi_2$ is linear to the auxiliary variable $u_{k}^\mathcal{P}$ thanks to $\dot \theta_k^\mathcal{P}$ depicting a linear system w.r.t. $u_{k}^\mathcal{P}$ as shown in \eqref{eqn:linear}.
Combining (\ref{eqn:function2}) with (\ref{eqn:linear}), 
the following theorem shows the invariance:
\begin{theorem}\label{thm:inva}
Given a neural ODE defined by (\ref{eqn:NN}) and an output specification $h(\bm x) \geq 0$, 
if there exist class $\mathcal{K}$ functions $\alpha_1, \alpha_2$ and $u_k^\mathcal{P}$ such that with $\psi_1,\psi_2$ as in \eqref{eqn:function2}, 
\begin{equation} \label{eqn:inva}
\begin{aligned}
 \Psi(u_k^\mathcal{P}|\bm x) = \psi_2(\bm x, u_k^\mathcal{P}) \geq 0,
    \end{aligned}
\end{equation}
for all $\bm x$ that satisfies $h(\bm x)\geq 0$ and $\psi_1(\bm x)\geq 0$, where $\Psi(u_k^\mathcal{P}|\bm x) =  \frac{d^2h(\bm x)}{d \bm x ^2} f_{\theta}^2(\bm x) + \frac{dh(\bm x)}{d \bm x}\frac{\partial f_{\theta}(\bm x)}{\partial \theta_{k}^\mathcal{P}}(A_{k}^\mathcal{P} \theta_{k}^\mathcal{P} + B_{k}^\mathcal{P} u_{k}^\mathcal{P}) + (\frac{dh(\bm x)}{d \bm x}\frac{\partial f_{\theta}(\bm x)}{\partial \bm x} + \frac{d\alpha_1(h(\bm x))}{d \bm x})f_{\theta}(\bm x) + \alpha_2(\psi_1(\bm x))$,
then the neural ODE is forward invariant.
\end{theorem}
The proof and existence of $\alpha_1, \alpha_2$ are shown in Appendix \ref{asec:proof2}. Intuitively, invariance is first propagated via $\psi_1$ to $\theta_k^\mathcal{P}$, then via $\psi_2$ to $u_k^\mathcal{P}$, rendering a linear constraint in \eqref{eqn:inva}. We will further show how this enforces invariance in Sec. \ref{sec:a2}.

\textbf{Brief Summary.} Note that (\ref{eqn:inva}) is linear in $u_k^\mathcal{P}$ with the assistance of system (\ref{eqn:linear}).  Instead of directly changing the parameters of the neural ODE for the invariance as Sec. \ref{sec:output}, we find auxiliary control $u_k^\mathcal{P}$ that satisfies the constraint (\ref{eqn:inva}) to dynamically change the parameters. Also, since we take the derivative of $h(\bm x)$ twice, as in (\ref{eqn:inva}), this is analogous to a second-order HOCBF (i.e., $m = 2$ in Def. \ref{def:hocbf})

 \textbf{IP to the Proper Parameters.} While Thm. \ref{thm:inva} allows us to propagate the invariance to arbitrary neural ODE parameters, the choice of the parameters may affect the performance, e.g., the model's accuracy). The specific parameter choice depends on the model structure and the task's output specification. 
 In most cases, we may wish to choose the parameters of the same layer to propagate the invariance to. However, it is also possible to choose parameters of different layers, as long as we define auxiliary dynamics for all the parameters as in (\ref{eqn:linear}). The proposed method still works in such cases. We may need to choose the parameters such that the output of the neural ODE can all be changed, as in the linear case.

 \subsection{Invariance Propagation to External Input}

We consider a neural ODE in the form of \eqref{eqn:NN_control} with an external input $\textbf{I}$ . 
 
\textbf{Approach 1:} As in Sec. \ref{sec:output}, we may directly reformulate (\ref{eqn:NN_control}) in the following affine form:
 \begin{equation}
    \dot {\bm x} = f_{\theta}(\bm x) + g_{\theta}(\bm x)\textbf{I},
\end{equation}
where $f_{\theta}$ is defined as in (\ref{eqn:NN}), $g_{\theta}:\mathbb{R}^n \rightarrow\mathbb{R}^{n \times n_\textbf{I}}$ is another neural network parameterized by $\theta$. Then, we can use the similar technique as in Sec. \ref{sec:output} to propagate the invariance to the external input $\textbf{I}$ as they are both in affine forms.

\textbf{Approach 2:} If we do wish to keep neural ODEs with external input as in the form of (\ref{eqn:NN_control}), then we may define auxiliary linear dynamics as in  Sec. \ref{sec:param}, and augment (\ref{eqn:NN_control}) by the following form:
\begin{equation}
\begin{aligned}
\dot {\bm x} = f_{\theta}^{'}(\bm x, \bm y), \qquad
\dot{\bm y} = A\bm y + B\textbf{I},
\end{aligned}
\end{equation}
where $\bm y\in\mathbb{R}^{n_\textbf{I}}$ is the auxiliary variable, $A\in\mathbb{R}^{n_\textbf{I}\times n_\textbf{I}}, B\in\mathbb{R}^{n_\textbf{I}\times n_\textbf{I}}$ are defined such that the linear system is controllable (similar to \eqref{eqn:linear}). Then, we can use the similar technique as in Sec. \ref{sec:param} to propagate the invariance to the external input $\textbf{I}$ via the auxiliary variable $\bm y$. In fact, the above neural ODE becomes a stacked neural ODE, which will be further studied (as discussed in Appendix \ref{asec:stacked}).

\section{Enforcing Invariance in Neural ODEs}
Here, we show how we proceed from the theoretical framework in Sec. \ref{sec:ip} to efficient algorithms of IP on neural ODEs.

\subsection{Algorithms for Linear layers}
\label{sec:a1}
 \textbf{Enforcing invariance.} Enforcing the invariance of a neural ODE is equivalent to the satisfaction of the condition in Thm.~\ref{thm:inv}. Also by proof in Appendix \ref{asec:proof1}, we can always find a class $\mathcal{K}$ function $\alpha_1(\cdot)$ such that there exists $\theta_{K-1,K}^\mathcal{P}$ that makes (\ref{eqn:inv}) satisfied if $h(\bm x(t_0))\geq 0$. If $h(x_1(t_0)) \leq 0$, then the output of the neural ODE will be driven to satisfy $h(\bm x) \geq 0$ when the constraint (\ref{eqn:inv}) in Thm. \ref{thm:inv} is satisfied due to its Lyapunov property \cite{Aaron2012}. The enforcing of the invariance could vary in different applications and we do not restrict to exact methods. We provide a minimum-deviation quadratic program (QP) approach.

Let $\theta_{K-1,K}^{\mathcal{P}\dagger}\in\mathbb{R}^{n\times n_{K-1}^\mathcal{P}}$ denote the 
 value of $\theta_{K-1,K}^\mathcal{P}$ during training or after training. Then, we can formulate the following optimization:
 \begin{equation} \label{eqn:qp}
     \theta_{K-1,K}^{\mathcal{P}*} = \arg\min_{\theta_{K-1,K}^{\mathcal{P}}} ||\theta_{K,K-1}^\mathcal{P} - \theta_{K-1,K}^{\mathcal{P}\dagger}||^2, \text{ s.t. } (\ref{eqn:inv}),
 \end{equation}
 where $||\cdot||$ denotes the Euclidean norm. The above optimization becomes a QP with all other variables fixed except $\theta_{K-1,K}^\mathcal{P}$. This solving method has been shown to work in \citep{Ames2017} \citep{Glotfelter2017} \citep{Xiao2021TAC2}. At each discretization step, we solve the above QP and get $\theta_{K-1,K}^{\mathcal{P}*}$. Then we set $\theta_{K-1,K}^{\mathcal{P}} = \theta_{K-1,K}^{\mathcal{P}*}$ during the inference of the neural ODE. This way, we can enforce the invariance, i.e., guarantee that $h(\bm x(t))\geq 0, \forall t\geq t_0$.
The process is summarized in Algorithm \ref{alg:pp}.

\textbf{Complexity of Enforcing Invariance.} The computational complexity of the QP (\ref{eqn:qp}) is $\mathcal{O}(q^3)$, where $q = n_{K-1}^\mathcal{P}n$.  When there is a set $S$ of output specifications, we just add the corresponding constraint (\ref{eqn:inv}) for each specification to  (\ref{eqn:qp}), and the number of constraints will not significantly increase the complexity. It is also possible to get the closed-form solution of the QP \citep{Ames2017} when there are only a few output specifications.

 \subsection{Algorithms for Nonlinear layers}
\label{sec:a2}
\textbf{Stability of Auxiliary Systems.} In this case, we need to make sure that the parameter $\theta_{k}^\mathcal{P}$ of the neural ODE is stabilized as it is dynamically controlled by (\ref{eqn:linear}). To enforce this, we use control Lyapunov functions (CLFs) \cite{Aaron2012}. Specifically, for each $\theta_{k_j}^\mathcal{P}, j\in\{1\dots, d_k^\mathcal{P}\}$, where $\theta_{k_j}^\mathcal{P}$ is a component of $\theta_{k}^\mathcal{P}$, we define a CLF $V(\theta_{k_j}^\mathcal{P}) = (\theta_{k_j}^\mathcal{P} - \theta_{k_j}^{\mathcal{P}\dagger})^2$, where $\theta_{k_j}^{\mathcal{P}\dagger}$ is the value of $\theta_{k_j}^\mathcal{P}$ during or after training. Then, any $u_k^\mathcal{P}$ that satisfies:
 \begin{equation} \label{eqn:inv_stab}
     \Phi(u_k^\mathcal{P}|\theta_{k_j}^\mathcal{P}) \leq 0, j\in\{1,\dots,d_k^\mathcal{P}\},
 \end{equation} 
 where $\Phi(u_k^\mathcal{P}|\theta_{k_j}^\mathcal{P}) = \frac{dV(\theta_{k_j}^\mathcal{P})}{d\theta_{k_j}^\mathcal{P}} (A_{k_j}^\mathcal{P} \theta_{k}^\mathcal{P} + B_{k_j}^\mathcal{P} u_{k}^\mathcal{P}) + \epsilon_j V(\theta_{k_j}^\mathcal{P})$, $A_{k_j}^\mathcal{P}\in\mathbb{R}^{1\times d_k^\mathcal{P}}, B_{k_j}^\mathcal{P}\in\mathbb{R}^{1\times d_k^\mathcal{P}}$ are the $j$'th rows of $A_k^\mathcal{P}, B_k^\mathcal{P}$ in (\ref{eqn:linear}), respectively and $\epsilon_j > 0$, will render the auxiliary systems \eqref{eqn:linear} stable. The proof is in Appendix \ref{asec:proof3}.

\textbf{Enforcing invariance.} Enforcing the invariance of a neural ODE is equivalent to the satisfaction of the condition in Thm.~\ref{thm:inva}. By proof in Appendix \ref{asec:proof2}, we can always find class $\mathcal{K}$ functions $\alpha_1, \alpha_2$ such that there exists $u_k^\mathcal{P}$ that makes (\ref{eqn:inva}) satisfied if $h(\bm x(t_0))> 0$. Again, we provide a minimum-deviation quadratic program (QP) approach:
\begin{equation} \label{eqn:qpa}
\begin{aligned}
    (u_{k}^{\mathcal{P}*}, &\delta_{1:d_k^\mathcal{P}}^*) = \arg\min_{u_{k}^{\mathcal{P}}, \delta_{1:d_k^\mathcal{P}}} ||u_k^\mathcal{P}||^2 + \sum_{j = 1}^{d_k^\mathcal{P}}w_j\delta_j^2, \\
    & \text{s.t. (\ref{eqn:inva}) and } \Phi(u_k^\mathcal{P}|\theta_{k_j}^\mathcal{P})\leq \delta_j, j\in\{1,\dots,d_k^\mathcal{P}\},
\end{aligned}
\end{equation}
 where $\delta_j\in\mathbb{R}$ is a slack variable that makes the CLF constraint soft (not conflict with (\ref{eqn:inva})), and $w_j > 0, j\in\{1,\dots, d_k^\mathcal{P}\}$ are pre-defined coefficients of penalties on the relaxations. The above optimization becomes a QP with all other variables fixed except $u_{k}^\mathcal{P}, \delta_j$, as discussed at the end of the last subsection. At each discretization step, we solve the above QP and get $u_k^{\mathcal{P}*}$. Then, the optimal parameter $\theta_k^{\mathcal{P}^*}$ is determined by the integration of (\ref{eqn:linear}) with $u_k^\mathcal{P} = u_k^{\mathcal{P}*}$, and set $\theta_{k}^\mathcal{P} = \theta_k^{\mathcal{P}^*}$ during the inference. This way, we can enforce the invariance, i.e., guarantee that $h(\bm x(t))\geq 0, \forall t\geq t_0$. We summarize the process in Algorithm \ref{alg:pp}.

\begin{algorithm}[tb]
   \caption{Invariance Propagation to Parameters}
   \label{alg:pp}
\begin{algorithmic}
   \STATE {\bfseries Input:} Output specification set $S$, trained or in-training neural ODE (\ref{eqn:NN}).
   \STATE (a) Choose the model parameters $\theta_k^\mathcal{P}$ that we wish to propagate the invariances (from the set $S$) to.
   \STATE (b) Make the chosen parameters $\theta_k^\mathcal{P}$ as symbolic variables.
   \STATE (c) Make $f_{\theta}(\bm x)$ in the neural ODE (\ref{eqn:NN}) as a symbolic function in terms of $\theta_k^\mathcal{P}$ and $\bm x$.
   \IF{\textit{Propagate to nonlinear layers}}
   \STATE (d) Define controllable linear dynamics (\ref{eqn:linear}) for $\theta_k^\mathcal{P}$.
   \STATE (e) Propagate invariances to $u_k^\mathcal{P}$ of system (\ref{eqn:linear}) by (\ref{eqn:inva}).
   \STATE (f) Define CLFs to stabilize $\theta_k^\mathcal{P}$ by (\ref{eqn:inv_stab}).
   \STATE (g) Formulate the QP (\ref{eqn:qpa}).
   \ELSE
    \STATE (d) Formulate the QP (\ref{eqn:qp}).
   \ENDIF
   \REPEAT
   \STATE Get the trained or in-training value $\theta_k^{\mathcal{P}\dagger}$ of $\theta_k^\mathcal{P}$.
   \IF{\textit{Propagate to nonlinear layers}}
   \STATE Solve the QP (\ref{eqn:qpa}) and get $u_k^{\mathcal{P}*}$.
   \STATE Get $\theta_k^{\mathcal{P}*}$ by integrating (\ref{eqn:linear}) with $u_k^\mathcal{P} = u_k^{\mathcal{P}*}$ ($u_k^\mathcal{P}$ is piecewise constant).
   \ELSE
   \STATE Get $\theta_k^{\mathcal{P}*}$ by solving the QP (\ref{eqn:qp}).
   \ENDIF
   \STATE Set $\theta_k^\mathcal{P} = \theta_k^{\mathcal{P}*}$ for the neural ODE (\ref{eqn:NN}).
   \UNTIL{$Training$ or $Inference$ is $done$}
\end{algorithmic}
\end{algorithm}

\textbf{Complexity of enforcing invariance} The computational complexity of the QP (\ref{eqn:qpa}) is $\mathcal{O}(q^3)$, where $q = 2d_k^\mathcal{P}$.  When there is a set $S$ of output specifications, we just add the corresponding constraint (\ref{eqn:inva}) for each specification to  (\ref{eqn:qpa}). The complexity is a little higher than the one in the case of invariance enforcement in a linear layer. This is due to the fact that the dimension of decision variables is doubled. Nonetheless, this is still efficient to solve.

\subsection{Training Neural ODEs with Invariance.} 
The invariance of neural ODEs can be enforced even after the training of neural ODEs. However, we need to hand-tune the parameters of class $\mathcal{K}$ functions in the QP (\ref{eqn:qp}) or (\ref{eqn:qpa}) to address the conservativeness of this approach, which is non-trivial when we have many output specifications. 

We leverage the power of differentiable QP \citep{Amos2017}. \textit{For neural ODEs whose invariance  are enforced on the external input} $\textbf{I}$, the differentiable QP that enforces the invariance is stacked to the neural ODE, and thus, the training is performed via the standard stochastic gradient descent. \textit{For neural ODEs whose invariance are enforced on the model parameter} $\theta$ ($\theta$ is either $\theta_{K-1,K}^\mathcal{P}$ as in Sec. \ref{sec:a1} or $\theta_{k}^\mathcal{P}$ as in Sec. \ref{sec:a2}), it is challenging to train both the neural ODE and differentiable QP simultaneously in the same pipeline. Thus, we propose the following two-stage training method: In the first stage, we train the neural ODE as usual and thus optimize the weight $\theta^{\dagger}$ of the network.  In the second stage, we train via the differentiable QP (more precisely, the parameter of class $\mathcal{K}$ functions in QP \eqref{eqn:qp}, \eqref{eqn:qpa}) such that $\theta$ minimally deviates from $\theta^{\dagger}$. The training of the network and the QP can be performed alternatively. We summarize the training process in Fig.~\ref{fig:training} in Appendix \ref{asec:train}.

\section{Experiments}
We set up experiments to answer the following questions:
\vspace{-2ex}
\begin{itemize}
[align=right,itemindent=0em,labelsep=2pt,labelwidth=1em,leftmargin=*,itemsep=0em] 

    \item Does our algorithm match the theoretical potential in various learning tasks both quantitatively and qualitatively?
    \item How does our invariance propagation compare with state-of-the-art approaches for enforcing output specifications?
    \item How does our proposed method scale with the number of parameters of the neural ODE and handle complex specifications of dynamical systems? 
\end{itemize}


\subsection{Spiral Curve Regression with Specifications}
In this experiment, we aim to impose trajectory constraints on the spiral dynamics task proposed in \citep{chen2018neural}.
The training data comes from solving an ODE $[\dot x, \dot y]^T = A[x^3, y^3]^T$, where $A = [-0.1, -2.0; 2.0, -0.1]$. We use a neural ODE to fit the data. We additionally require the trajectory $\bm x = (x,y)$ to avoid some areas defined by $h_j(\bm x) \geq 0, j\in S$, where $S$ denotes a set of constraints. 

\textbf{Comparison between invariance enforcing on different layers.} We first compare the performance of neural ODEs when propagating invariances to different layers and different numbers of model parameters. As shown in Table \ref{tab:layer}, The output specifications are all guaranteed (Sat. $\geq 0$) with invariance propagation, which shows the flexibility of the method. While the output specifications are violated in pure neural ODEs. The computation time is higher when we propagate the invariances to the hidden layer than one of the output layers, and the computation time slightly increases when we significantly increase the number of chosen parameters, although the inference errors do not actually vary too much. An illustrative example is shown in Figs. \ref{fig:layers}-\ref{fig:sat} in Appendix \ref{asec:spiral}.

\begin{table}[ht]
\caption{Spiral curve comparisons between neural ODE and invariance propagated  to different layers. The numbers in brackets denote the number of parameters chosen in the invariance. Sat. (Satisfaction) $\geq 0$ denotes the satisfaction of output specifications. }
\label{tab:layer}
\vskip 0.05in
\begin{center}
\begin{small}
\resizebox{.47\textwidth}{!}{
\begin{sc}
\begin{tabular}{lcccr}
\toprule
Method &  Sat. ($\geq 0$) & MSE ($\downarrow$) & time ($\downarrow$) \\
\midrule
neural ODE   & -0.031 & 0.510 & 0.003 \\
hidden inv. (6) & 0.003& 0.393 & 0.025\\
hidden inv. (20)    & 0.001 &0.391 & 0.030 \\
hidden inv. (60)   & $4e^{-4}$ & 0.387 & 0.044  \\
hidden inv. (100)     & $9e^{-4}$& 0.391 & 0.057\\
output inv. (6)      & $2e^{-4}$ & 0.441 & 0.007 \\
output inv. (20)      & $2e^{-4}$& 0.442 &    0.010   \\
output inv. (60)   & $2e^{-4}$ & 0.442 & 0.017 \\
output inv. (100)   & $2e^{-4}$ & 0.442 & 0.025 \\
\bottomrule
\end{tabular}
\end{sc}
}
\end{small}
\end{center}
\vskip -0.1in
\end{table}

\begin{figure}[t]
	\centering
	\includegraphics[width=\columnwidth]{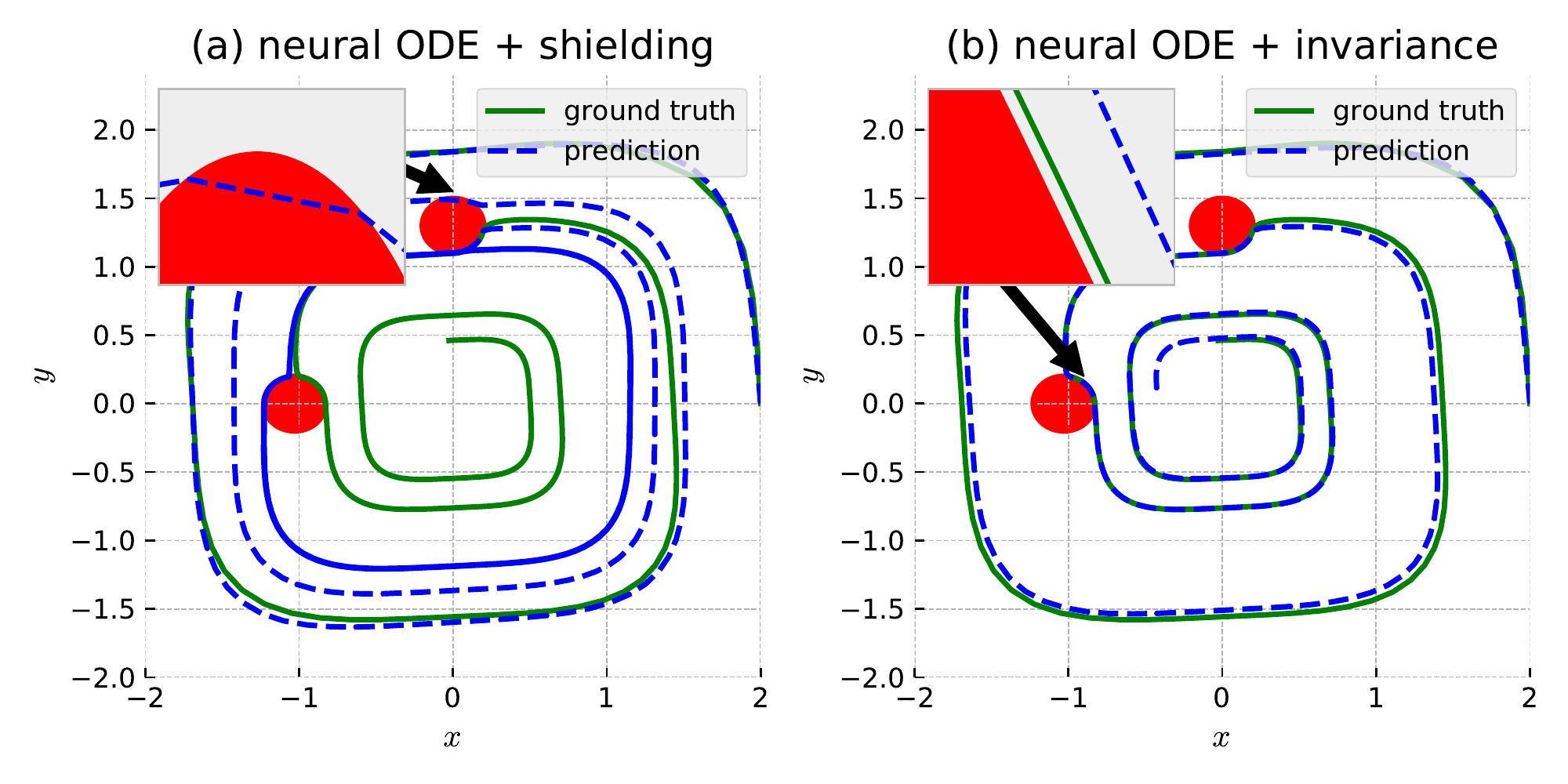} 
	\caption{Invariance in training for spiral curve regression with output specifications.  The shielding method can only guarantee point-wise satisfaction, and thus the specification can still be violated between samplings. }%
	\label{fig:spiral_tr}%
\end{figure}

\textbf{Comparison with benchmarks.} The proposed invariance allows us to enforce arbitrary specifications after training with a different number of parameters. An illustrative example is shown in Fig. \ref{fig:spiral} in Appendix \ref{asec:spiral}.
 With invariance in the training loop, the model outputs can strictly satisfy the output specifications while staying close to the ground truth (see Figure~\ref{fig:spiral_tr}b).  The comparisons between our proposed (output) invariance with other benchmarks are shown in Table \ref{tab:comp_sp} in which the results are evaluated using 100 trained models for each method.  Compared to the shielding method \citep{Ferlez2020}, our invariance model can achieve better performance. Most importantly, the proposed invariance can guarantee more complex specifications including addressing the inter-sampling effect, i.e., specification satisfaction between sampling time, as shown in Fig. \ref{fig:spiral_tr}. Both the filter approach \citep{pereira2020} and BarrierNet \citep{Xiao2021bnet} perform badly when they are placed outside the neural ODE (less computationally expensive than the case when they are placed inside the neural ODE). Moreover, the filter approach is not trainable, which may introduce the worse performance. In contrast, the performance of our proposed invariance is similar when the QP is placed inside and outside the neural ODE (we only show the result when the QP is placed inside the neural ODE  in Table \ref{tab:comp_sp}), which shows its flexibility.

 \begin{table}[ht]
\caption{Spiral curve comparisons with benchmarks. [I]/[O] denotes the Filter \citep{pereira2020} or BarrierNet (BNet) \citep{Xiao2021bnet}  is inside/outside the neural ODE \citep{chen2018neural}.  Items are short for In-loop training test mean-squared error (IN-LOOP MSE), Post-training test mean-squared error (POST MSE), Complex specifications and Inter-sampling effect (COMP. \& IS), Pointwise guarantee (PW GUAR.), respectively. Shielding is \citep{Ferlez2020}. The method items are short for Neural ODE (NEUR. ODE), Neural ODE with low conservative training (NODE-L), Neural ODE with high conservative training (NODE-H), respectively.}
\label{tab:comp_sp}
\vskip 0.05in
\begin{center}
\begin{small}
\resizebox{.47\textwidth}{!}{
\begin{sc}
\begin{tabular}{p{1.8cm}<{\centering}p{1.3cm}<{\centering}<{\centering}p{1.3cm}<{\centering}p{0.8cm}<{\centering}p{0.8cm}<{\centering}r}
\toprule
Method & In-loop MSE($\downarrow$)  & Post MSE($\downarrow$)     & Comp. \& IS    & PW guar.  \\
\midrule
Neur. ODE
& $0.49 {\small \pm 0.17}$ & $0.49{\small\pm 0.17}$   &$\times$  & $\times$ \\
NODE-L
& $0.59 {\small \pm 0.18}$ & $0.59 {\small \pm 0.18}$   &$\times$  & $\times$ \\
NODE-H
& $0.73 {\small \pm 0.18}$ & $0.73 {\small \pm 0.18}$   &$\times$  & $\times$ \\
\midrule
Shielding
     & $0.61{\small\pm 0.12}$ & $0.61{\small\pm 0.10}$  &$\times$   &  $\surd$    \\
     
Filter [I]
& N/A & $0.55{\small\pm 0.21}$  &$\surd$   &  $\surd$    \\

Filter [O]
& N/A & $1.27{\small\pm 0.14}$  &$\surd$   &  $\surd$    \\

  BNet [I]
  & $0.64 {\small\pm 0.09}$ & $0.44{\small\pm 0.08}$  &$\surd$   &  $\surd$    \\
  BNet [O]
  & N/A & $1.04{\small\pm 0.11}$  &$\surd$   &  $\surd$    \\
  Inv. (Ours)  & $0.55{\small\pm 0.12}$ &$0.44{\small\pm 0.08}$  & $\surd$&  $\surd$
\\\bottomrule
\end{tabular}
\end{sc}
}
\end{small}
\end{center}
\vskip -0.1in
\end{table}

		
     



\subsection{Convexity Portrait of a Function} 
In this experiment, we assess whether our method can enforce that the neural ODE outputs satisfy Jensen's inequality. Jensen's inequality can be used to characterize whether a function is convex or not. In other words, a function $g$ is convex if the Jensen's inequality is satisfied: $
    \mu_1g(x) + \mu_2g(y) \geq g(\mu_1 x + \mu_2 y),
$
where $\mu_1\in[0,1], \mu_2\in[0,1]$ such that $\mu_1 + \mu_2 = 1$. A neural ODE is not guaranteed to satisfy Jensen's inequality as illustrated by the red-dashed curve in Figure~\ref{fig:convex_in}b (in appendix). However, with the proposed (hidden and output) invariances, the model outputs are guaranteed to satisfy Jensen's inequality, as shown by the blue-dashed and cyan dashed curves in Figure~\ref{fig:convex_in}b (in appendix).

\subsection{HalfCheetah-v2 and Walker2d-v2} 
\begin{figure}[t]
    \centering
    \subfigure{\includegraphics[width=0.45\linewidth]{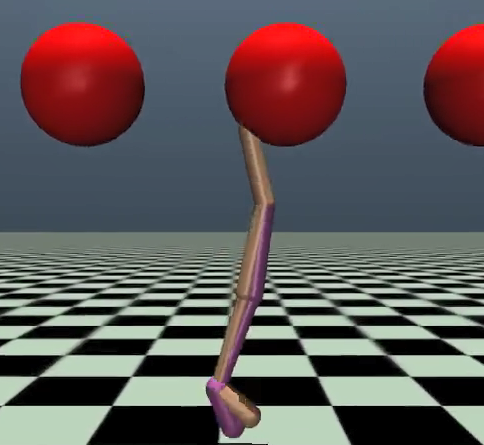}}
     \subfigure{\includegraphics[width=0.45\linewidth]{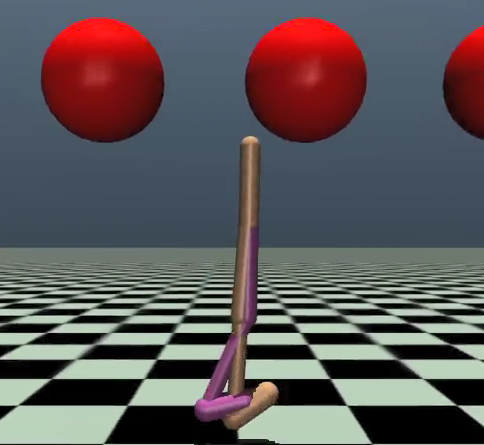}}
    
    \caption{Planning for Walker2d-v2 in obstacle avoidance using neural ODE (left) and invariance (right). Collisions happen when using neural ODE only, while they can be avoided with guarantees using the proposed invariance method.
    }
    \label{fig:walker_cbf}
\end{figure}

In this section, we evaluate our invariance framework on two publicly available datasets for modeling physical dynamical systems \cite{lechner2022mixed,hasani2021liquid}.
The two datasets consist of trajectories of the HalfCheetah-v2 and Walker2d-v2 3D robot systems \cite{gym} generated by the Mujoco physics engine \cite{todorov2012mujoco}. Each trajectory represents a sequence of a 17-dimensional vector describing the system's state, such as the robot's joint angles and poses. For each of the two tasks, we define 34 safety constraints that restrict the system's evolution to the value ranges observed in the dataset. We compare our invariance approach with a hard truncation of the system state, i.e., projecting points violating the constraints to the nearest points that satisfy them. Our invariance framework can achieve competitive performance compared to other approaches, as shown in Table \ref{tab:comp_h2}, while guaranteeing the satisfaction of complex safety specifications. We enforced our invariance on 17, 34, and 170 model parameters, respectively. We also present a case study  (Fig. \ref{fig:walker_cbf}) in planning for obstacle avoidance for the Walker2d-v2 in which the truncation method fails to work.

\begin{table}[ht]
\caption{Walker2d-v2 and halfcheetah-v2 comparisons between Neural ODE \citep{chen2018neural}, Truncation \cite{gym}, and Invariance (ours). Items are short for Walker2d-v2 test mean-squared error (W2D MSE), halfcheetah-v2 test mean-squared error (HC MSE), Complex specifications and Inter-sampling effect (COMP. \& IS), Joint limit satisfaction (SAFETY), respectively.}
\label{tab:comp_h2}
\vskip 0.05in
\begin{center}
\begin{small}
\resizebox{.47\textwidth}{!}{
\begin{sc}
\begin{tabular}{p{1.8cm}<{\centering}p{1.3cm}<{\centering}<{\centering}p{1.3cm}<{\centering}p{0.8cm}<{\centering}p{0.9cm}<{\centering}r}
\toprule
Method & \makecell[t]{W2d\\ MSE ($\downarrow$)}  & \makecell[t]{HC\\ MSE ($\downarrow$)}    &  Comp. \& IS  & Safety ($\geq 0$)  \\
\midrule
Neur. ODE & $1.06{\small\pm 0.07}$ & $2.17{\small\pm 0.03}$    &$\times$  & -1.78      \\
		\midrule
	    Truncation  & $1.15{\small\pm 0.08}$ & $2.17{\small\pm 0.03}$   &$\times$   &  -8.13     \\
		Inv. (Ours)  & $1.06{\small\pm 0.07}$ &$2.13{\small\pm 0.02}$  & $\surd$&  0.0   \\\bottomrule
\end{tabular}
\end{sc}
}
\end{small}
\end{center}
\vskip -0.1in
\end{table}

		
		

\subsection{Lidar-based End-to-End Autonomous Driving}

In this section, we consider Lidar-based end-to-end autonomous driving that has complex specifications from dynamics. The approach of finding neural ODE specifications from dynamics can be found in Appendix \ref{asec:complex}. 
The neural ODE takes a Lidar point cloud as input $\textbf{I}$, and outputs controls for the autonomous vehicle to follow the lane. The problem and training setup is shown 
 in Appendix \ref{asec:driving}.
 
With noisy Lidar, the neural ODE controller may cause the ego vehicle to collide with the other moving vehicle during the overtaking process (the red trajectory shown in Figure~\ref{fig:snap} in Appendix \ref{asec:driving}).  Although with safety guarantees, the resulting trajectory from a safe filter \citep{pereira2020} may make the ego vehicle conservative (as the blue trajectory shown in Figure~\ref{fig:snap}), and thus stay unnecessarily far away from the optimal trajectory (ground truth). The BarrierNet \citep{Xiao2021bnet} is also a filter, but it addresses conservativeness by including the CBF (filter) in the training loop. However, the training of a BarrierNet is harder compared with the invariance as reference controls and the relative weight among them should also be trained in addition to the CBF parameters.
 We summarize this comparison in Table \ref{tab:comp} that includes testing results of 100 case studies under noisy lidar perception. The invariance has the least conservativeness while guaranteeing safety.

\begin{table}[ht]
\caption{Self-driving comparisons between safe filter \citep{pereira2020}, BarrierNet \citep{Xiao2021bnet}, neural ODE \citep{chen2018neural} and invariance (ours). Items are short for Trajectory test  mean-squared error (TRAJ. MSE), Conservativeness measurement (CONSER.), Safety measurement (SAFETY), Model complexity (MOD. CMP.), respectively. L, H are shorts for Low and High.}
\vspace{-3mm}
\label{tab:comp}
\begin{center}
\begin{small}
\resizebox{.49\textwidth}{!}{
\begin{sc}
\begin{tabular}{p{1.7cm}<{\centering}p{1.4cm}<{\centering}<{\centering}p{1.5cm}<{\centering}p{1.0cm}<{\centering}p{0.4cm}<{\centering}p{0.5cm}<{\centering}r}
\toprule
Method & Traj. MSE ($\downarrow$) & Conser. ($\geq0$ \& $\downarrow$) & Safety ($\geq 0$)   & Dy. free   & Mod. cmp. \\
\midrule
Neur. ODE &$0.46{\small\pm 0.04}$ & $-13.1{\small\pm 1.49}$ & $-17.26$ &$\surd$ & l   \\
		
		Safe filter     & $0.96{\small\pm 0.04}$ & $27.69{\small\pm 1.08}$& 24.60 & $\times$  &h   \\
		
		BarrierNet  & $0.34{\small\pm 0.01}$& $8.51{\small\pm 0.33}$ & 7.69 &$\times$ & h  \\
		
		Inv. (Ours)  & $0.36{\small\pm 0.01}$ & $1.97{\small\pm 0.06}$ & 1.83  &$\surd$  &l \\\bottomrule
\end{tabular}
\end{sc}
}
\end{small}
\end{center}
\vskip -0.2in
\end{table}

		
		
		
		

 \section{Related Works}

\noindent \textbf{Neural ODEs for imitation learning.} Neural ODEs \citep{chen2018neural} \citep{chen2020learning} are powerful dynamical systems modeling tools, widely used in applications to learning system kinetics \citep{kim2021stiff} \citep{alvarez2020dynode} \citep{baker2022learning}, in graphics \citep{PhysRevResearch.4.013221}, in discovering novel materials \citep{chen2022forecasting}, and in robot controls \citep{hasani2017compositional,amini2020learning,lechner2022mixed,lechner2020neural,vorbach2021causal}. Neural ODEs are continuous-time universal approximators \citep{kidger2020neural} that perform competitive to their static and discretized neural network counterparts, once their complexity issues \citep{massaroli2020dissecting} are resolved by better numerical solvers \citep{poli2020hypersolvers}, or by their closed-form variants \citep{hasani2022closed}. Recent methods provide safety guarantees for inference in a neural ODE system, e.g. stochastic reachability analysis \citep{gruenbacher2020verification}. However, there are no methods to simultaneously train the model while guaranteeing safety. Here, we address this issue by forward-invariance of neural ODEs. 

\noindent \textbf{Set invariance and CBFs.} An invariant set has been widely used to characterize the safe behavior of dynamical systems \citep{preindl2016robust} \citep{rakovic2005invariant} \citep{Ames2017} \citep{Glotfelter2017} \cite{Xiao2021bnet}. In the state of the art, Control Barrier Functions (CBFs) are also widely used to prove set
invariance \citep{Aubin2009}, \citep{Prajna2007}, \citep{Wisniewski2013}. They can be traced back to optimization problems
\citep{Boyd2004}, and are Lyapunov-like functions \citep{Tee2009}, 
\citep{Wieland2007}. Existing CBF approaches have significant limitations: They fail on systems with unknown dynamics, provide rather conservative guarantees, and work efficiently only for affine systems. Our work addresses all these limitations.

\noindent \textbf{Existing approaches for guarantees in neural networks.}
Recent advances in differentiable optimization methods show promise for safety-guaranteed neural network controllers \citep{pereira2020, Amos2018, Xiao2021bnet, wang2023learning}. The differentiable optimizations are usually served as a layer (filter) in the neural networks. In \citep{Amos2017}, a differentiable quadratic program (QP) layer, called OptNet, was introduced. OptNet with CBFs has been used in neural networks as a filter for safe controls \citep{pereira2020}, in which CBFs are not trainable, thus, potentially limiting the system's learning performance.  In \citep{Deshmukh2019,Jin2020,Zhao2021, Ferlez2020}, safety guaranteed neural network controllers have been learned through verification-in-the-loop training.  The verification approaches cannot ensure coverage of the entire state space. 
While the proposed invariance can avoid such issues, and generalize to a wide class of guarantees.

 \section{Conclusions, Discussions and Future Work}
We have demonstrated the effectiveness of our invariance propagation method in ensuring the safe operation of neural ODE instances in a series of dynamical system learning tasks. Nonetheless, our method faces a few shortcomings which provide directions for future work to focus on.

In particular, our method requires neural ODEs to have continuously differentiable activation functions. 
Moreover, our method requires prior knowledge of output specifications. 
Future work may investigate how to learn specifications from observational data for unknown specifications. Finally, we observed the propagation of invariance is subjected to the vanishing gradient for deeper networks. We can alleviate this shortcoming by gradient preservation methods such as mixed-memory ODE-based networks \cite{lechner2022mixed} which we will investigate in future work. 


\section*{Acknowledgements}

The research was supported in part by Capgemini Engineering. It was also partially sponsored by the United States Air Force Research Laboratory and the United States Air Force Artificial Intelligence Accelerator and was accomplished under Cooperative Agreement Number FA8750-19-2-1000. The views and conclusions contained in this document are those of the authors and should not be interpreted as representing the official policies, either expressed or implied, of the United States Air Force or the U.S. Government. The U.S. Government is authorized to reproduce and distribute reprints for Government purposes notwithstanding any copyright notation herein. This research was also supported in part by the AI2050 program at Schmidt Futures (Grant G-
965 22-63172),


\bibliography{MCBF}
\bibliographystyle{icml2023}

\newpage
\appendix
\onecolumn

\section{Proof}

\subsection{Proof of Theorem \ref{thm:inv} (Invariance Propagation to Linear Layers)}
\label{asec:proof1}
 Given a continuously differentiable constraint $h(x)\geq 0$ ($h(\bm x(t_0))\geq 0$), by Nagumo's theorem \citep{Nagumo1942berDL}, the necessary and sufficient condition for the satisfaction of $h(x(t))\geq 0, \forall t\geq t_0$ is 
 $$\dot h(x) \geq 0, \text{ when } h(x) = 0$$

The neural ODE is reformulated as
$$\dot {\bm x} = \sum_{i=1}^{n_2}\theta_{K-1,K}^i {\bm z}_{K-1}^i = \theta_{K-1,K}^{\mathcal{P}}{\bm z}_{K-1}^{\mathcal{P}} + \theta_{K-1,K}^{\mathcal{N}}{\bm z}_{K-1}^{\mathcal{N}}
$$
and the condition in the theorem is given by
$$\Psi(\theta_{K-1,K}^{\mathcal{P}}|\bm x) = \underbrace{\frac{dh(\bm x)}{d \bm x}\theta_{K-1,K}^{\mathcal{P}}{\bm z}_{K-1}^{\mathcal{P}} + \frac{dh(\bm x)}{d \bm x}(\theta_{K-1,K}^{\mathcal{N}}{\bm z}_{K-1}^{\mathcal{N}})}_{L_{f_{\theta}}h(\bm x)} + \alpha_1(h(\bm x)) \geq 0.$$
Where $f_{\theta}(\bm x) = \theta_{K-1,K}^{\mathcal{P}}{\bm z}_{K-1}^{\mathcal{P}} + \theta_{K-1,K}^{\mathcal{N}}{\bm z}_{K-1}^{\mathcal{N}}$.
Combining the last two equations, we have 
$$\Psi(\theta_{K-1,K}^{\mathcal{P}}|\bm x) = \frac{dh(\bm x)}{d \bm x}\dot {\bm x} + \alpha_1(h(\bm x)) \geq 0.$$
which is equivalent to
$$\psi_1(\bm x, \theta_{K-1,K}^{\mathcal{P}}) = \dot h(\bm x) + \alpha_{1}(h(\bm x)) \geq 0,$$ 
 Since $\alpha_{1}$ is a class $\mathcal{K}$ function, $$\alpha_{1}(h(\bm x)) \rightarrow 0 \text{ as } h(\bm x)\rightarrow 0.$$
In other words, we have $\dot h(\bm x)\geq 0$ when $h(\bm x)= 0$. Then, by Nagumo's theorem, we have that $h(\bm x) \geq 0$ is satisfied if $h(\bm x(t_0))\geq 0$ since 
the derivative of $h(\bm x)$ is non-decreasing on the hyperplane $h(\bm x) = 0$. In other words, we have that $$h(\bm x(t))\geq 0, \forall t\geq 0,$$ and the neural ODE is invariant. $\hfill\blacksquare$

\textbf{Existence of class $\mathcal{K}$ function $\alpha_1$ in Theorem \ref{thm:inv}:}

Given a continuously differentiable $h(\bm x)\geq 0$. If $h(\bm x(t_0))\geq 0$, there always exists such a class $\mathcal{K}$ function $\alpha_1$ in (\ref{eqn:inv}) that allows us to freely choose $\theta_{K-1,K}^{\mathcal{P}}$ to make (\ref{eqn:inv}) satisfied as $\theta_{K-1,K}^{\mathcal{P}}$ is usually unconstrained. $\hfill\blacksquare$

\subsection{Proof of Theorem \ref{thm:inva} (Invariance Propagation to Nonlienar Layers)}
\label{asec:proof2}

The auxiliary dynamics are defined as
$$
\dot \theta_{k}^\mathcal{P} = A_{k}^\mathcal{P} \theta_{k}^\mathcal{P} + B_{k}^\mathcal{P} u_{k}^\mathcal{P}
$$

The condition in the theorem is given as
$$\Psi(u_k^\mathcal{P}|\bm x) =  \underbrace{\frac{d^2h(\bm x)}{d \bm x ^2} f_{\theta}^2(\bm x) + (\frac{dh(\bm x)}{d \bm x}\frac{\partial f_{\theta}(\bm x)}{\partial \bm x} + \frac{d\alpha_1(h(\bm x))}{d \bm x})f_{\theta}(\bm x)}_{L_{f_{\theta}}^2h(\bm x) + L_{f_{\theta}} (\alpha_1\circ h)(\bm x)} + \underbrace{\frac{dh(\bm x)}{d \bm x}\frac{\partial f_{\theta}(\bm x)}{\partial \theta_{k}^\mathcal{P}}(A_{k}^\mathcal{P} \theta_{k}^\mathcal{P} + B_{k}^\mathcal{P} u_{k}^\mathcal{P})}_{L_{g_u}L_{f_{\theta}}h(\bm x)}  + \alpha_2(\psi_1(\bm x)) \geq 0,$$
Where $g_u = A_{k}^\mathcal{P} \theta_{k}^\mathcal{P} + B_{k}^\mathcal{P} u_{k}^\mathcal{P}$.
Combining the last two equations, we have
$$\Psi(u_k^\mathcal{P}|\bm x) =  \frac{d^2h(\bm x)}{d \bm x ^2} f_{\theta}^2(\bm x) + \frac{dh(\bm x)}{d \bm x}\dot f_{\theta}(\bm x) + \frac{d\alpha_1(h(\bm x))}{d t} + \alpha_2(\psi_1(\bm x)) \geq 0,$$

Since $\psi_1(\bm x) = \dot h(\bm x) + \alpha_{1}(h(\bm x)),$
the above equation 
 is equivalent to 
$$
\psi_2(\bm x) :=\dot \psi_1(\bm x)  + \alpha_2(\psi_1(\bm x)) \geq 0,
$$ where $\dot \psi_1$ is involved with the derivative of $\theta_k^p$ that is defined by the auxiliary dynamics. 
Since $\bm x(t_0)$ is such that $\psi_1(\bm x(t_0))\geq 0$, then by Theorem. \ref{thm:inv}, we have that 
$$\psi_1(\bm x(t))\geq 0, \forall t\geq t_0.$$

Further, $$\psi_1(\bm x) = \dot h(\bm x) + \alpha_{1}(h(\bm x)) \geq 0,$$ following  (\ref{eqn:function1}). Again by Theorem. \ref{thm:inv}, since $h(\bm x(t_0))\geq 0$, we have that $$h(\bm x(t))\geq 0, \forall t\geq t_0,$$ and thus the neural ODE (\ref{eqn:NN}) is invariant. $\hfill\blacksquare$

Using a similar technique as in the proof of Theorem \ref{thm:inva}, the proposed invariance is provably correct for stacked neural ODEs that may introduce higher relative degree $h(\bm x)$.

\textbf{Existence of class $\mathcal{K}$ functions $\alpha_1, \alpha_2$ in Theorem \ref{thm:inv}:}

Given a continuously differentiable $h(\bm x)\geq 0$. If $h(\bm x(t_0))> 0$, there always exists such a class $\mathcal{K}$ function $\alpha_1$ such that $\psi_1(\bm x(t_0)) > 0$ as $\psi_1(\bm x) = \dot h(\bm x) + \alpha_{1}(h(\bm x))$. Then, we can always find a class $\mathcal{K}$ function $\alpha_2$ in (\ref{eqn:inva}) that allows us to freely choose $u_{k}^p$ to make (\ref{eqn:inva}) satisfied as $u_{k}^p$ is usually unconstrained. 

On the other hand, if $h(\bm x(t_0)) = 0$ and $\dot h(\bm x(t_0)) \geq 0$, we can also find such class $\mathcal{K}$ functions $\alpha_1, \alpha_2$ in (\ref{eqn:inva}) following similar analysis.

Finally, if $h(\bm x(t_0)) = 0$ and $\dot h(\bm x(t_0)) < 0$, then the specification will be immediately violated following Nagumo's theorem, and there is no way to enforce invariance. $\hfill\blacksquare$

\subsection{Proof of Stability of Auxiliary Systems}
\label{asec:proof3}
The auxiliary dynamics are defined as
$$
\dot \theta_{k}^\mathcal{P} = A_{k}^\mathcal{P} \theta_{k}^\mathcal{P} + B_{k}^\mathcal{P} u_{k}^\mathcal{P}
$$

The CLF constraint in the invariance enforcing algorithm in the nonlinear case (the relaxation variable $\delta_j = 0$ when the CLF constraint does not conflict with the invariance enforcing constraint, i.e., when the output (state) of the neural ODE is far from undesired set boundaries) is 
$$\Phi(u_k^\mathcal{P}|\theta_{k_j}^\mathcal{P}) = \frac{dV(\theta_{k_j}^\mathcal{P})}{d\theta_{k_j}^\mathcal{P}} (A_{k_j}^\mathcal{P} \theta_{k}^\mathcal{P} + B_{k_j}^\mathcal{P} u_{k}^\mathcal{P}) + \epsilon_j V(\theta_{k_j}^\mathcal{P}) \leq 0, j\in\{1,\dots, d_k^\mathcal{P}\},$$

Combining the last two equations, we have 
\begin{equation}\label{eqn:clf1}
\Phi(u_k^\mathcal{P}|\theta_{k_j}^\mathcal{P}) = \dot V(\theta_{k_j}^\mathcal{P}) + \epsilon_j V(\theta_{k_j}^\mathcal{P}) \leq 0, j\in\{1,\dots, d_k^\mathcal{P}\},
\end{equation}

Suppose we have 
$$\dot V(\theta_{k_j}^\mathcal{P}) + \epsilon_j V(\theta_{k_j}^\mathcal{P}) = 0, $$
the solution to the above equation is
$$V(\theta_{k_j}^\mathcal{P}(t)) =  V(\theta_{k_j}^\mathcal{P}(t_0))e^{-\epsilon_j (t-t_0)},$$
Using the comparison lemma \cite{Khalil2002}, equation (\ref{eqn:clf1}) implies that
$$
V(\theta_{k_j}^\mathcal{P}(t)) \leq  V(\theta_{k_j}^\mathcal{P}(t_0))e^{-\epsilon_j (t-t_0)}, j\in\{1,\dots, d_k^\mathcal{P}\},
$$
Therefore,
$$
V(\theta_{k_j}^\mathcal{P}(t)) \rightarrow 0, \text{ as } t\rightarrow \infty, \forall j\in\{1,\dots, d_k^\mathcal{P}\},
$$
and $\theta_{k_j}^\mathcal{P}$ is exponentially stabilized to $\theta_{k_j}^\mathcal{P\dagger}$ as $V(\theta_{k_j}^\mathcal{P}) = (\theta_{k_j}^\mathcal{P} - \theta_{k_j}^\mathcal{P\dagger})^2$. $\hfill \blacksquare$

\newpage
\section{Discussion on Auxiliary Dynamics}
\label{asec:aux}
The auxiliary dynamics are defined as
\begin{equation}
    \dot \theta_{k}^\mathcal{P} = A_{k}^\mathcal{P} \theta_{k}^\mathcal{P} + B_{k}^\mathcal{P} u_{k}^\mathcal{P}
\end{equation}
where the above system is controllable if $A_{k}^\mathcal{P}$ and $B_{k}^\mathcal{P}$ are chosen such that 
$$
\left[B_{k}^\mathcal{P}, A_{k}^\mathcal{P}B_{k}^\mathcal{P}, \dots,  (A_{k}^\mathcal{P})^{m-1}B_{k}^\mathcal{P}\right]
$$
is in full rank, where $m\in\mathbb{N}$ is a finite time step that drives the system from an initial state to a final state.

The exact choice of $A_{k}^\mathcal{P}$ and $B_{k}^\mathcal{P}$ may affect the performance. In other words, they will determine how close the output trajectory of the neural ODE can stay from the boundaries of undesired sets, and they will also determine how the parameters vary. From our experience, the effect on performance is pretty small since both the auxiliary control $u_k^\mathcal{P}$ and the parameter $\theta_k^\mathcal{P}$ are unbounded in the neural ODE, and thus, we can always quickly change the parameter $\theta_k^\mathcal{P}$ under any $A_{k}^\mathcal{P}$ and $B_{k}^\mathcal{P}$ that make the auxiliary system controllable. For simplicity, we can choose $A_{k}^\mathcal{P}$ and $B_{k}^\mathcal{P}$ to be zero and identity matrices, respectively.

On the other hand, we can make $A_{k}^\mathcal{P}$ and $B_{k}^\mathcal{P}$ be trainable parameters in our QP implementation via differentiable QP \citep{Amos2017}.

\newpage
\section{Training neural ODEs with Invariance}
\label{asec:train}
The main objective is to make sure neural ODEs satisfy specifications during training, which also allows us to train class $\mathcal{K}$ functions in Theorems \ref{thm:inv} and \ref{thm:inva}.

\begin{figure}[ht]
\begin{center}
\centerline{\includegraphics[width=0.7\columnwidth]{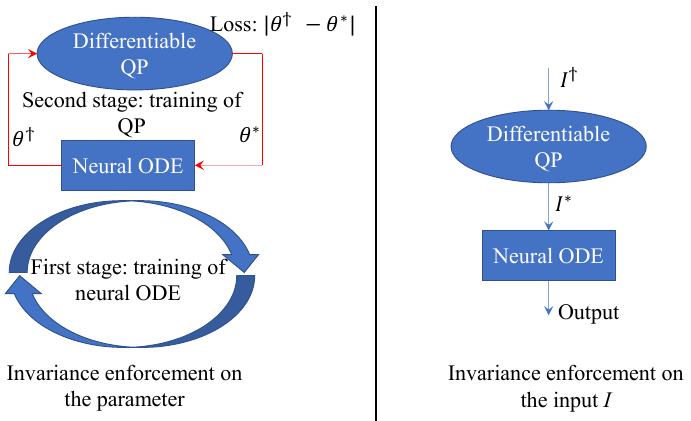}}
\caption{Training neural ODEs with invariance. The training of a neural ODE with $\textbf{I}$ is regular when we enforce the invariance on $\textbf{I}$. }
\label{fig:training}
\end{center}
\end{figure}

For neural ODEs that enforce the invariance on the input $\textbf{I}$, the training is performed via the standard stochastic gradient descent. While training neural ODEs for enforcing the invariance on the model parameter $\theta$, it is challenging to train both the neural ODE and differentiable QP simultaneously in the same pipeline. Thus, we propose the following two-stage training method: In the first stage, we train the neural ODE as usual and thus optimize the weight $\theta^{\dagger}$ of the network.  In the second stage, we train the differentiable QP (specifically, the parameter in class $\mathcal{K}$ functions in QP (\ref{eqn:qp})) such that $\theta$ minimally deviates from $\theta^{\dagger}$. The training of the network and the QP can be performed alternatively. We summarize the training process in Fig.~\ref{fig:training}.

\newpage
\section{Complex Specifications}
\label{asec:complex}
Although the invariance of neural ODEs can be applied to a wide class of problems, one of the important applications is in the safe control of dynamical systems, as this involves complex output specifications. Consider the case where the output of the neural ODE controller is directly taken as the control of the dynamical system. The dynamical system is usually required to satisfy some safety constraints that are defined over its state instead of over its control. In other words, the specification of the neural ODE is not directly on its output. To map a state constraint onto the control of a dynamics system (i.e., the output of the neural ODE), we can use the CBF method.

More specifically, consider a control system whose dynamics are defined in the form:
\begin{equation} \label{eqn:dynamics}
    \dot {\bm y} = f(\bm y, \bm u),
\end{equation}
where $\bm y\in \mathbb{R}^{q}$ is the state of the system, $\bm u = \bm x$ is its control (i.e., the output of the neural ODE). $f:\mathbb{R}^{q\times n_1}\rightarrow \mathbb{R}^q$, where $n_1$ is the dimension of $x_1$ (or the control).

We wish the state of the system (\ref{eqn:dynamics}) to satisfy the following (safety) constraint:
\begin{equation}\label{eqn:safety}
    b(\bm y) \geq 0,
\end{equation}
where $b:\mathbb{R}^q\rightarrow \mathbb{R}$ is continuously differentiable, and its relative degree with respect to $\bm u$ is $d\in\mathbb{N}$.

As shown in \citep{Xiao2021TAC2}, we can use a HOCBF to enforce the safety constraint (\ref{eqn:safety}) for system (\ref{eqn:dynamics}). In other words, we map the safety constraint (\ref{eqn:safety}) onto the following HOCBF constraint:
\begin{equation} \label{eqn:hocbf}
   \frac{d\phi_{d-1}}{d\bm y}f(\bm y, \bm u) + \alpha_d(\phi_{d-1}(\bm y)) \geq 0,
\end{equation}
where $\phi_k(\bm y) = \dot \phi_{k-1}(\bm y) + \alpha_k(\phi_{k-1}(\bm y)), k = \{1,\dots, d-1\}$ and $\phi_0(\bm y) = b(\bm y)$, $\alpha_k, k\in\{1,\dots,d\}$ are class $\mathcal{K}$ functions. It is worth noting  that the construction of the HOCBF is similar to the construction (\ref{eqn:function1}) of the invariance of a neural ODE. The satisfaction of the above HOCBF constraint (\ref{eqn:hocbf}) implies the satisfaction of the safety constraint (\ref{eqn:safety}).

Since the output of the neural ODE is used to control the dynamical system, i.e., $\bm u = \bm x$, we can find the output constraint of the neural ODE from (\ref{eqn:hocbf}) in the form:
\begin{equation} \label{eqn:out_c}
   h(\bm x, \bm y) = \frac{d\phi_{d-1}}{d\bm y}f(\bm y, \bm x) + \alpha_d(\phi_{d-1}(\bm y)) \geq 0,
\end{equation}
Then, we can back-propagate the invariance of the neural ODE (i.e., $h(\bm x(t), \bm y(t))\geq 0, \forall t\geq 0$) to the input $\textbf{I}$ or its parameter, as shown before.
In fact, in this scenario, the neural ODE serves as an integral controller for dynamical systems. In the original CBF method, we need to assume that the dynamics (\ref{eqn:dynamics}) are in affine control form in order to use the CBF-based QP to efficiently find a safe controller. With the proposed method, such restriction (assumption) is removed. This shows the advantage of the invariance of neural ODEs in safety-critical control problems.

\newpage
\section{Stacked Neural ODEs} 
\label{asec:stacked}
When we have stacked neural ODEs, the IP is similar. Suppose we wish to enforce the invariance on the $k_{th}$ stacked neural ODE, then we may choose the $k_{th}$ stacked neural ODE to have linear layers, and we can get a similar equation as (\ref{eqn:NN_affine}). The difference is that we may need to define higher-order CBFs. The performance of stacked neural ODEs needs to be further studied before applying the proposed invariance. Note that the index of stacked neural ODEs is in the reverse order as the layer index in non-stacked neural ODEs to make a difference between them.

\begin{figure}[ht]
	\centering
	\includegraphics[scale=1.6]{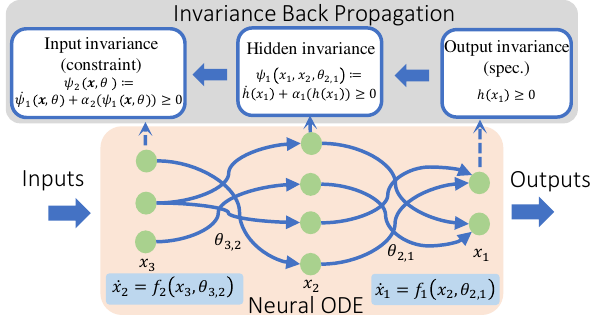} 
	\caption{Relative degree and invariance example of three stacked neural ODEs. Recurrence is allowed, i.e., $f_1, f_2$ {\color{black}(neural networks)} could be a function of $x_1, x_2$, respectively. $x_3$ is the input of the stacked neural ODE. {\color{black}There is an \textit{odeint} after each layer}. }%
	\label{fig:example}%
\end{figure}

For example, in Figure~\ref{fig:example}, the outputs of the neurons in $x_3$ are the inputs of the neurons in $x_2$. The neurons in $x_3$ are with one relative degree higher than the ones in $x_2$. The same applies to the neurons of $x_2$ and $x_1$. The highest relative degree of the three-layer neural ODE is two since the output neurons $x_1$ are defined to be with a relative degree of 0. The output specification $h(\bm x)\geq 0$ of a neural ODE can then be rewritten as $h(x_1) \geq 0$, as $x_1$ is the vector of output neurons. In order to show the relationships between the invariances of different layers of a neural ODE,  we define the first-from-the-last hidden layer invariance $\psi_1 \geq 0$ (defined similarly as in Definition~\ref{def:inv}) as a function of $h(x_1)$ and its derivative, where $\psi_1$ is defined as:
\begin{equation}
    \psi_1(x_1, x_2, \theta_{2,1}) := \dot h(x_1) + \alpha_1(h(x_1)),
\end{equation}
where $\theta_{2,1}$ is the connection weight between layers 2 and 1, $x_2, \theta_{2,1}$ shows up in $\dot h(x_1)$, and $\alpha_1(\cdot)$ is a class $\mathcal{K}$ function (a class $\mathcal{K}$ is a strictly increasing function that passes through the origin). This way, the hidden invariance is related to the neurons $x_1, x_2$ and their connection weight $\theta_{2,1}$. We can define the invariance of any hidden (or input) layer $k$ by functions $\psi_{k-1}
(x_1, \dots, x_{k}, \theta_{2,1}, \dots,\theta_{k,k-1})
\geq 0, k \in\{2, \dots,  m+1\}$,  recursively:
\begin{equation} \label{eqn:if}
\begin{aligned}
    &\psi_{k-1}(x_1, \dots, x_{k}, \theta_{2,1}, \dots,\theta_{k,k-1})  := \dot\psi_{k-2}(x_1, \dots, x_{k-1},\theta_{2,1}, \dots,\theta_{k-1,k-2}) \\&\qquad+ \alpha_{k-1}(\psi_{k-2}(x_1, \dots, x_{k-1},\theta_{2,1}, \dots,\theta_{k-1,k-2})), \quad k\in\{1,\dots, m+1\},
\end{aligned}
\end{equation}
where $\alpha_{k-1}, k \in\{2, \dots, m + 1\}$ are class $\mathcal{K}$ functions, and $\psi_0(x_1, \theta_{1,0}) = h(x_1)$.  $\theta_{k,k-1}$ denotes the connection weight between layers $k$ and $k-1$. For the input layer $m + 1$, we have $\psi_m(\bm x, \theta) = \psi_m(x_1, \dots, x_{m+1}, \theta_{2,1}, \dots, \theta_{m+1,m})$, where $\theta = (\theta_{2,1}, \dots, \theta_{m+1,m})$, and the invariance of the input layer is illustrated by the input constraint $\psi_m(\bm x, \theta) \geq 0$. For example,  the highest relative degree of the three stacked neural ODEs in Figure~\ref{fig:example} is 2, and the input constraint (invariance) is $\psi_2(x_1, x_2, x_3, \theta_{2,1}, \theta_{3,2}) \geq 0$. Again, the performance of stacked neural ODEs should be studied first compared to non-stacked neural ODEs, we leave this for future work.

\textbf{Order of HOCBFs.} The order of HOCBFs in IP equals the relative degree of the specification (if it is from a dynamic system) plus the number of stacked neural ODEs, and plus one (if enforced on a nonlinear layer).

\newpage
\section{Experiment Details}
In this section, we provide detailed settings for all the experiments, including some additional figures and results.

\subsection{Spiral Curve Regression with Specifications}
\label{asec:spiral}

In this case, we enforce invariance on parameters in both the hidden nonlinear layer and the output linear layer.

\textbf{Training data generation.} The initial condition for the ODE we sample the data from is $[2, 0]$, and we sampled 1000 data points within the time interval [0,25] as the training data set. In order to make sure that the sampled data avoids the two critical regions in the case of invariance-in-training, we use CBFs to minimally modify the ODE. In other words, the components $A[0,1], A[1,0]$ of the A matrix in the considered ODE are minimally changed by the following quadratic program:
\begin{equation}
    \min_{a_1, a_2} (a_1 - A[0,1])^2 + (a_2 - A[1,0])^2
\end{equation}
s.t. CBF constraints:
\begin{equation}
\begin{aligned}
    2(A[0,1] - O_{x,1})A[1,0]^3a_1 +  2(A[1,0] - O_{y,1})A[0,1]^3a_2 + h_1(x) \geq 0,\\
    2(A[0,1] - O_{x,2})A[1,0]^3a_1 +  2(A[1,0] - O_{y,2})A[0,1]^3a_2 + h_2(x) \geq 0,
    \end{aligned}
\end{equation}
where $h_i(x) = (A[0,1] - O_{x,i})^2 + (A[1,0] - O_{y,i})^2 - R_i^2, i\in\{1,2\}$, and $(o_{x,i}, o_{y,i})$ denotes the location of the undesired set $i$, $R_i$ denotes its size ($R_1 = R_2 = 0.2$ in the experiments).

After solving the above QP at each time and obtaining $a_1^*, a_2^*$, we replace $A[0,1], A[1,0]$ by $a_1^*, a_2^*$, respectively, in the ODE (please find details in the attached code).

\textbf{Model structure.} 
The training implementation and the enforcing QP for the invariance of the neural ODE are also given in the attached code.  The $f_{\theta}$ in the neural ODE (\ref{eqn:NN}) is a three-layer fully connected network with sizes 2, 50, and 2, respectively. The activation functions used in the hidden layers are tanhshrink, while the output layers are without activation functions (linear layers).

\textbf{Training.} The training epoch is 500, and the training batch size is 20 with a batch sequence time of 10. We use RMSprop optimizer with learning rate $1e^{-3}$. The training time is about 2 hours on an RTX3090 GPU.

\textbf{Invariance enforcing on the output linear layer.}  In this case, the neural ODE can be rewritten in linear form as in (\ref{eqn:NN_affine}) in terms of the output layer parameters. Therefore, the invariance implementation can be achieved by directly changing the output layer parameters.

The specifications are defined as $h_i(\bm x) = (x - O_{x,i})^2 + (y - O_{y,i})^2 - R_i^2, i\in\{1,2\}$. Therefore, in (\ref{eqn:inv}), we have
$$\frac{dh_i(\bm x)}{d \bm x} = [2(x - O_{x,i}), 2(y - O_{y,i})],$$
and we choose the class $\mathcal{K}$ function $\alpha_1$ as a linear function.

The QP (\ref{eqn:qp}) in this case is
\begin{equation} 
     \theta_{K-1,K}^{\mathcal{P}*} = \arg\min_{\theta_{K-1,K}^{\mathcal{P}}} ||\theta_{K-1,K}^{\mathcal{P}} - \theta_{K-1,K}^{\mathcal{P}\dagger}||^2,
 \end{equation}
 \text{ s.t. }
 \begin{equation}
\frac{dh_1}{d \bm x}\theta_{K-1,K}^{\mathcal{P}}{\bm z}_{K-1}^{\mathcal{P}} + \frac{dh_1}{d \bm x}\theta_{K-1,K}^{\mathcal{N}}{\bm z}_{K-1}^{\mathcal{N}}  + k_1h_1(\bm x) \geq 0,
\end{equation}
 \begin{equation}
\frac{dh_2}{d \bm x}\theta_{K-1,K}^{\mathcal{P}}{\bm z}_{K-1}^{\mathcal{P}} + \frac{dh_2}{d \bm x}\theta_{K-1,K}^{\mathcal{N}}{\bm z}_{K-1}^{\mathcal{N}} + k_2h_2(\bm x) \geq 0,
\end{equation}
where $k1 = k2 = 10$ when the invariance is enforced after the training of the neural ODE.

\textbf{Invariance enforcing on the hidden nonlinear layer.} 
The neural ODE (\ref{eqn:NN}) can be explicitly written as
\begin{equation} \label{eqn:NN_spiral}
    \dot {\bm x} = f_{\theta}(\bm x) =  \theta_{2,3}(tanhshrink(\theta_{1,2} \bm x^3 + b_{1,2})) + b_{2,3},
\end{equation}

The auxiliary dynamics (\ref{eqn:linear}) is defined as
\begin{equation} \label{eqn:aux}
    \dot \theta_{1}^\mathcal{P} = u_{1}^\mathcal{P},
\end{equation}
where $\theta_{1}^\mathcal{P}$ is the vector form of $\theta_{1,2}^\mathcal{P}$, and $\theta_{1,2}^\mathcal{P}$ is the partial parameter of $\theta_{1,2}$ in (\ref{eqn:NN_spiral}) that we wish to propagate the invariances to.

Therefore, in (\ref{eqn:inva}), we have
$$
\frac{dh_i(\bm x)}{d\bm x} = [2(x - O_{x,i}), 2(y - O_{y,i})], i\in\{1,2\}
$$
$$
\frac{\partial f_{\theta}(\bm x)}{\partial \theta_{1}^\mathcal{P}} = \theta_{2,3}tanh^2(\theta_{1,2} \bm x^3 + b_{1,2}) \dot\theta_{1,2} \bm x^3 (\text{part of } \dot\theta_{1,2} \text{ i.e., $\dot \theta_{1}^\mathcal{P}$, is defined as }u_1^\mathcal{P} \text{ following }(\ref{eqn:aux})),
$$
$$
\frac{\partial f_{\theta}(\bm x)}{\partial \bm x} = 3\theta_{2,3}tanh^2(\theta_{1,2} \bm x^3 + b_{1,2}) \theta_{1,2}\bm x^2 f_{\theta}(\bm x),
$$
$$
\frac{dh_i^2(\bm x)}{d\bm x^2} = [2, 2], i\in\{1,2\},
$$
and we choose class $\mathcal{K}$ functions $\alpha_1, \alpha_2$ as linear functions. The CLFs used to stabilize $\theta_1^\mathcal{P}$ is defined as (\ref{eqn:inv_stab}).

The QP (\ref{eqn:qpa}) in this case is
\begin{equation} 
    (u_{1}^{\mathcal{P}*}, \delta_{1:d_1^\mathcal{P}}^*) = \arg\min_{u_{1}^{\mathcal{P}}, \delta_{1:d_1^\mathcal{P}}} ||u_1^\mathcal{P}||^2 + \sum_{j = 1}^{d_1^\mathcal{P}}w_j\delta_j^2,
\end{equation}
\text{s.t. }
\begin{equation}
 \frac{d^2h_1(\bm x)}{d \bm x ^2} f_{\theta}^2(\bm x) + \frac{dh_1(\bm x)}{d \bm x}\frac{\partial f_{\theta}(\bm x)}{\partial \theta_{1}^\mathcal{P}}u_{1}^\mathcal{P} + (\frac{dh_1(\bm x)}{d \bm x}\frac{\partial f_{\theta}(\bm x)}{\partial \bm x} + \frac{dk_1h_1(\bm x)}{d \bm x})f_{\theta}(\bm x) + k_2\psi_{1,1}(\bm x) \geq 0,
\end{equation}
\begin{equation}
 \frac{d^2h_2(\bm x)}{d \bm x ^2} f_{\theta}^2(\bm x) + \frac{dh_2(\bm x)}{d \bm x}\frac{\partial f_{\theta}(\bm x)}{\partial \theta_{1}^\mathcal{P}} u_{1}^\mathcal{P} + (\frac{dh_2(\bm x)}{d \bm x}\frac{\partial f_{\theta}(\bm x)}{\partial \bm x} + \frac{dk_1h_2(\bm x)}{d \bm x})f_{\theta}(\bm x) + k_2\psi_{1,2}(\bm x) \geq 0,
\end{equation}
 \begin{equation} 
     \frac{dV(\theta_{1_j}^\mathcal{P})}{d\theta_{1_j}^\mathcal{P}} u_{1}^\mathcal{P} + \epsilon_j V(\theta_{1_j}^\mathcal{P}) \leq \delta_j, j\in \{1, \dots, d_1^\mathcal{P}\},
 \end{equation}
 where $k_1 = 20, k_2 = 100, \epsilon_j = 10, w_j = 1$ in Table \ref{tab:layer}. $\psi_{1,j}(\bm x) = \dot h_j(\bm x) + k_1h_j(\bm x), j\in\{1,2\}$.
 
\begin{figure}[ht]
\vskip 0.2in
\begin{center}
\centerline{\includegraphics[width=\columnwidth]{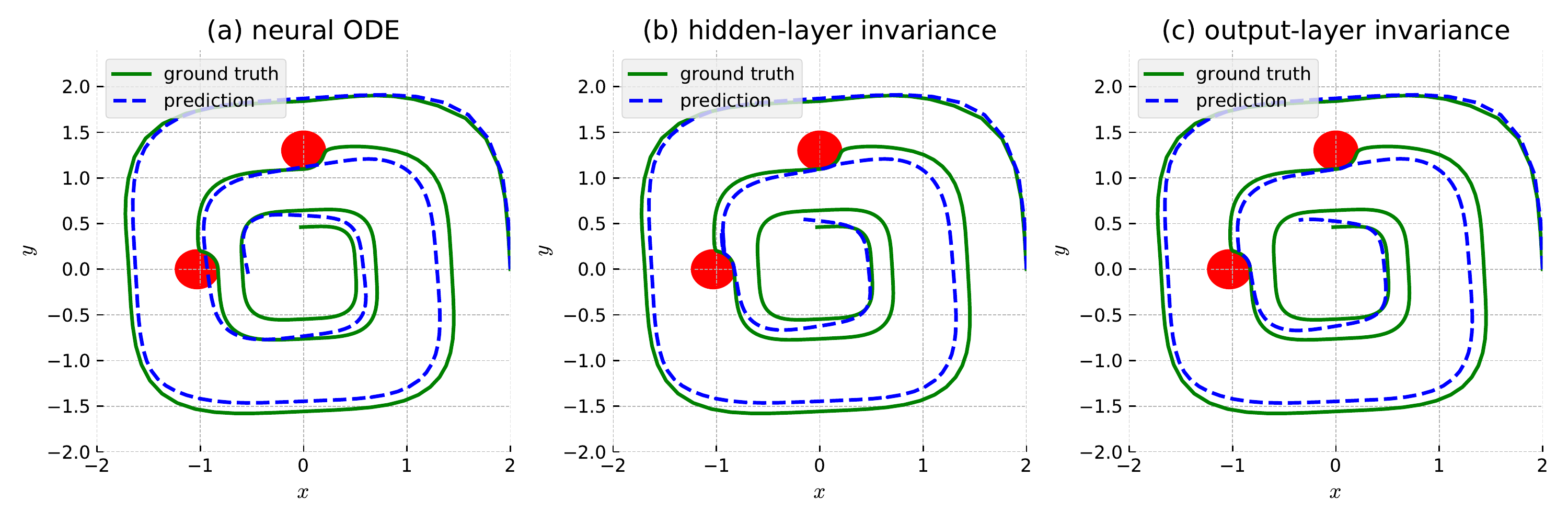}}
\caption{Spiral curve performance comparisons between neural ODE and invariance propagated to the hidden and output layers. The prediction of the neural ODE violates the output specifications. }
\label{fig:layers}
\end{center}
\vskip -0.2in
\end{figure}

\begin{figure}[ht]
\vskip 0.2in
\begin{center}
\centerline{\includegraphics[scale = 0.7]{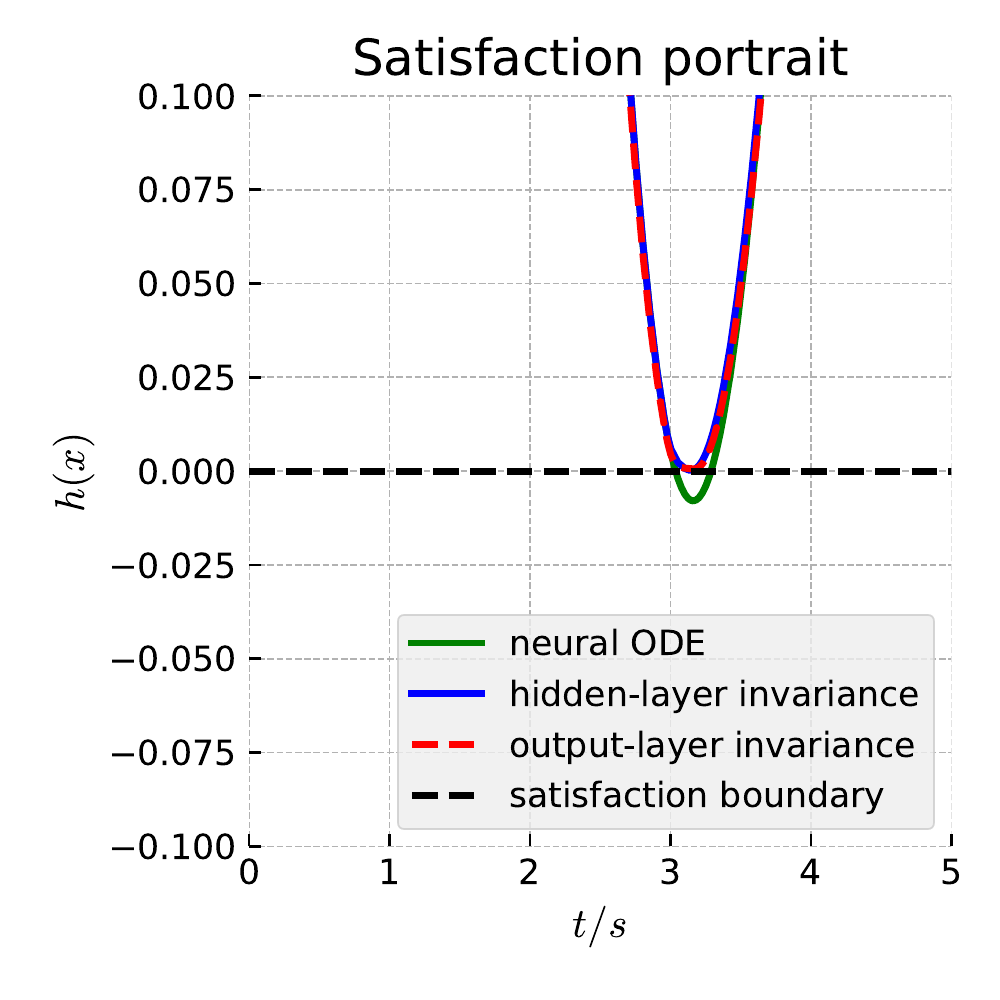}}
\caption{Spiral curve regression: specification satisfaction portraits. $h(\bm x) = \min\{h_1(\bm x), h_2(\bm x)\}\geq 0$} denotes the satisfaction of all the specifications. $h_j(\bm x) = (x - x_{o_j})^2 + (y - y_{o_j})^2 - R^2, j\in\{1,2\}$.
\label{fig:sat}
\end{center}
\vskip -0.2in
\end{figure}

\textbf{Comparison between invariance enforcing on different layers.} We present an illustrative example for the comparison between neural ODE and invariances enforced on different layers in Fig. \ref{fig:layers}. The chosen number of parameters is 6 for both the hidden invariance and output invariance. As expected, the output specifications are guaranteed to be satisfied in the invariances, while they are violated in the (pure) neural ODE.
We further illustrate the value of $h(\bm x) = \min\{h_1(\bm x), h_2(\bm x)\}$ in Fig. \ref{fig:sat}, in which $h(\bm x)\geq 0$ denotes the satisfaction of all the specifications.

\begin{figure}[t]
	\centering
	$\hspace{-6mm}$\includegraphics[scale=0.4]{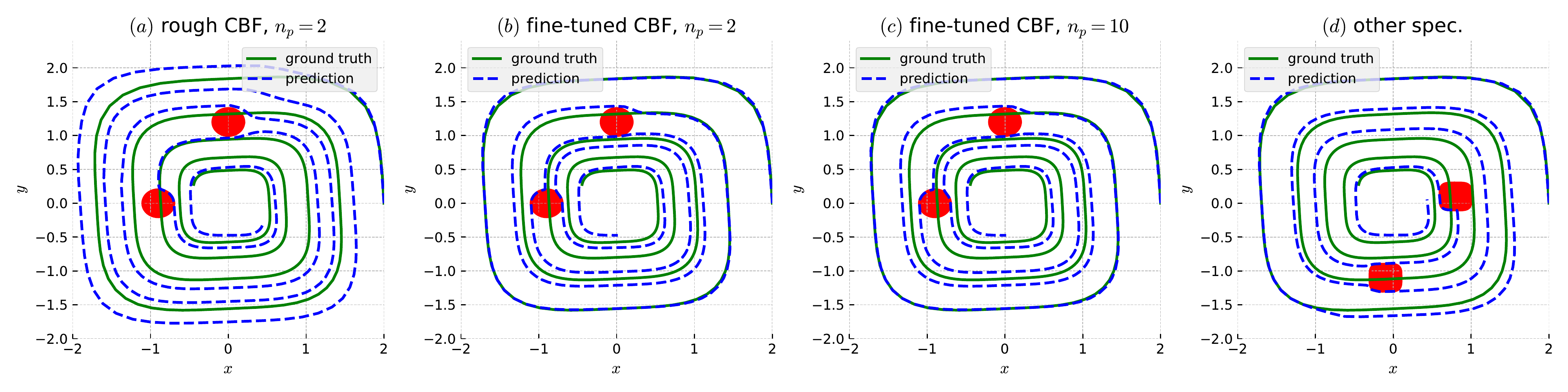}
	\vspace{-3mm}
	\caption{After-training invariance enforcing for spiral curve regression with output specifications. $n_p$ denotes the number of parameters randomly chosen in the QP (\ref{eqn:qp}). (a)-(c) are with two circular undesired sets defined by: $h_j(\bm x) = (x - x_{o_j})^2 + (y - y_{o_j})^2 - R^2$, ($x_{o_j}, y_{o_j}$) denotes the location of undesired set $j\in S, R > 0$, and (d) is with two randomly-placed superellipse-type undesired sets defined by: $h_j(\bm x) = (x - x_{o_j})^4 + (y - y_{o_j})^4 - R^4$.}%
	\label{fig:spiral}%
	\vspace{-1mm}
\end{figure}

\textbf{After training.}  In this case, since the neural ODE does not have the external input $\textbf{I}$, we minimally change the parameters of the model to enforce the invariance using the proposed QP-based approach (\ref{eqn:qp}). As a result, the outputs of the neural ODE can satisfy all the specifications (see Figures~\ref{fig:spiral} (a)-(d)). The model is already trained, we need to carefully choose the class $\mathcal{K}$ functions shown in (\ref{eqn:function1}).  Otherwise, the resulting trajectory would be overly-conservative such that there is a large deviation from the original one even when the state is far away from the constraint boundary $h_j(\bm x) = 0, j\in S$, as shown in Fig. \ref{fig:spiral}a. With slightly fined-tuned CBF parameters $p_i$ (i.e., its class $\mathcal{K}$ functions), the conservativeness can be addressed, as shown in Figure~\ref{fig:spiral}b.

We can also minimally change a different number of parameters of the neural ODE. Comparing Fig. \ref{fig:spiral}b with Figure~\ref{fig:spiral}c, the outputs of the neural ODE are almost the same under 2 or 10 randomly chosen model parameters. This demonstrates the effectiveness of the proposed invariance. Although the computation efficiency is better when we choose to modify fewer weights using the proposed QP-based approach (\ref{eqn:qp}), the performance might actually be worse when the learning task is complicated as we have to largely change the parameters. In order to show the robustness of the proposed invariance, we also tested other types of specifications, as the two randomly-placed superellipse-type undesired sets shown in Figure~\ref{fig:spiral}d. The outputs of the neural ODE can also guarantee the satisfaction of the corresponding output specifications. Tuning the CBF parameters in the invariance may be non-trivial when there are many specifications. Thus, we show next how we can enforce invariance in the training loop.

\newpage
\subsection{Convexity Portrait of a Function}
\label{asec:convex}
In this case, we enforce invariance on parameters in both the hidden nonlinear layer and the output linear layer.

\textbf{Training data generation.} The convex function we consider for sampling the training data is $g(x) = x^2$, and we set $\mu_1 = \mu_2 = 0.5$ for the Jensen's inequality. $x = t, y = t + 2 -\frac{1.9}{10}t$, where $t\in[0,10]$. We sampled 100 data points as the training data set. The CBF $h(x,y)$ for Jensen's inequality in the neural ODE is defined as:
\begin{equation}
    h(x,y) = \mu_1 z_1 + \mu_2 z_2 - z_3, 
\end{equation}
where $z_1, z_2, z_3$ denote the three outputs of the neural ODE. The implementation details and the enforcing QP for the invariance are given in the attached code.

\textbf{Model structure.} The $f_{\theta}$ in the neural ODE (\ref{eqn:NN}) is a three-layer fully connected network with sizes 3, 50, and 3, respectively. The activation functions used in the model are tanh.

\textbf{Training.} The training epoch is 2000, and the training batch size is 20 with a batch sequence time of 10. We use RMSprop optimizer with learning rate $1e^{-3}$. The training time is about 1 hour on an RTX3090 GPU.

\begin{figure}[ht]
	\centering
	\includegraphics[width=0.7\columnwidth]{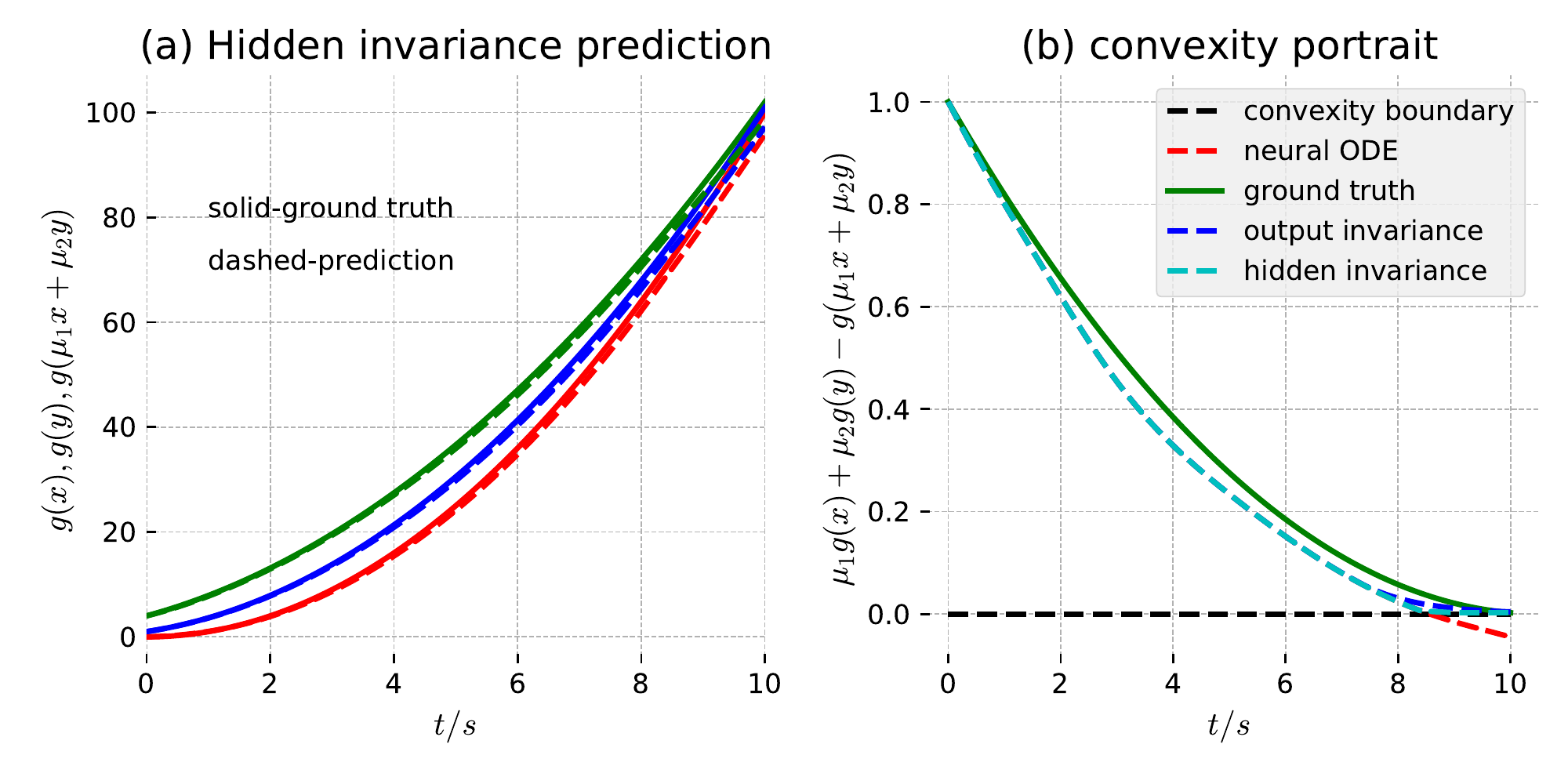} 
	\vspace{-2mm}\caption{Convexity portrait of the neural ODE outputs. The non-negativity of the functions in (b) demonstrates the satisfaction of Jensen's inequality. }%
	\label{fig:convex_in}%
\end{figure}

 \begin{figure}[ht]
	\centering
	$\hspace{-6mm}$\includegraphics[scale=0.45]{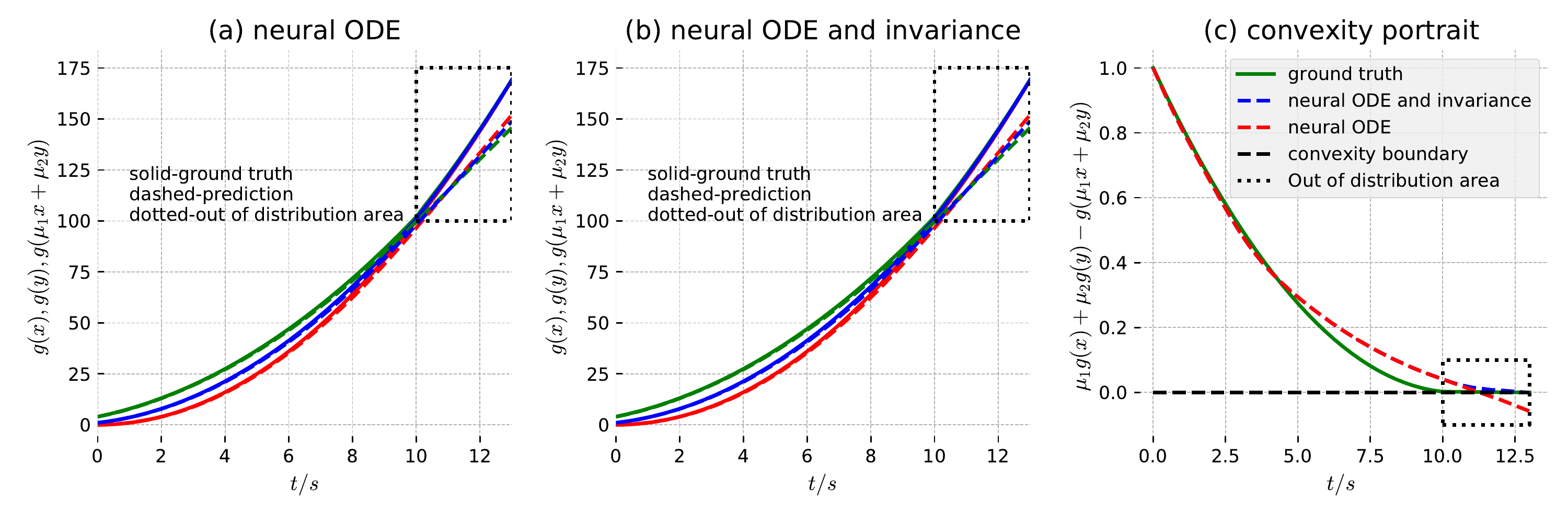} 
	\caption{Convexity portrait of the neural ODE outputs. $x, y$ are functions of $t$. 
	The non-negativity of the functions in (c) demonstrates the satisfaction of Jensen's inequality. }%
	\label{fig:convex_out}%
\end{figure}

\textbf{In distribution.} Within the range of the training data, the trained neural ODE is not guaranteed to satisfy Jensen's inequality as illustrated by the red-dashed curve in Figure~\ref{fig:convex_in}b. However, with the proposed (hidden and output) invariances, the model outputs are guaranteed to satisfy the Jensen's inequality, as shown by the blue-dashed and cyan dashed curves in Figure~\ref{fig:convex_in}b.

 \textbf{Out of distribution.} Although the outputs of a trained neural ODE model may satisfy Jensen's inequality within the range of the training data set, they may still violate Jensen's inequality when we conduct predictions for future time (out of the training data range), as shown by the red-dashed curve in Figure~\ref{fig:convex_out}c. 
However, with the proposed invariance, we can guarantee that the future prediction of the model also satisfies Jensen's inequality (see blue dashed line in Fig. \ref{fig:convex_out}c).

\newpage

\subsection{HalfCheetah-v2 and Walker2d-v2 kinematic modeling}
\label{asec:cheetah}
In this case, we enforce invariance on parameters in the output linear layer.

We evaluate our invariance framework on two publicly available datasets for modeling physical dynamical systems \cite{lechner2022mixed,hasani2021liquid}.
The two datasets consist of trajectories of the HalfCheetah-v2 and Walker2d-v2 3D robot systems \cite{gym} generated by the Mujoco physics engine \cite{todorov2012mujoco}. Each trajectory represents a sequence of a 17-dimensional vector describing the system's state, such as the robot's joint angles and poses. For each of the two tasks, we define 34 safety constraints that restrict the system's evolution to the value ranges observed in the dataset, i.e., the joint limitations.

\textbf{Model structure.} The $f_{\theta}$ in the neural ODE (\ref{eqn:NN}) is a three-layer fully connected network with sizes 17, 64, and 17, respectively. The activation functions used in the model are Tanh. 

\textbf{Training.} The training epoch is 200, and the training batch size is 64 with a batch sequence time of 20. We use RMSprop optimizer with learning rate $1e^{-3}$. The training time is about 1 hour on an RTX3090 GPU.

\newpage
\subsection{Lidar-based End-to-End Autonomous Driving}
\label{asec:driving}

In lidar-based driving, we assume 
the states of the ego and ado vehicles are obtained by other sensors (e.g. GPS or communication). We use the proposed invariance to back-propagate the safety requirements of the ego vehicle all the way to the input layer of the neural ODE, i.e., finding a constraint on the Lidar input $I$ that can guarantee the safety of the ego vehicle. 

\textbf{Problem setup.} The ego vehicle state $\bm x = (x, y, \theta, v)$ (along-lane location, off-center distance, heading, and speed, respectively) follows the unicycle vehicle dynamics, and the other vehicle moves at a constant speed. The ego vehicle is initially behind the other moving vehicle, and its \textit{objective} is to overtake the other vehicle while avoiding collisions. The collision avoidance is characterized by a safety constraint $b(\bm x, \bm x_p) = (x - x_p)^2 + (y - (y_p + y_d))^2 - R^2 \geq 0$, where $\bm x_p\in\mathbb{R}^4$ denotes the state of the preceding vehicle, and $(x_p, y_p)\in\mathbb{R}^2$ denotes the location of the preceding vehicle. $y_d\in\mathbb{R}$ is the off-center distance of the covering disk with respect to the center of the other vehicle. The satisfaction of $b(\bm x, \bm x_p)\geq 0$ implies collision-free. 

\textbf{Training setup.} In order to train a neural ODE controller that can be applied to the ego vehicle in closed-loop testing, 
we randomly assign locations for the ego vehicle with random states around the other vehicle. Then, we use a safety-guaranteed CBF-based QP controller to generate safe controls for the ego vehicle to overtake the other vehicle. We sampled 200 trajectories as the training data, and each trajectory has a time sequence of states and controls with a length of 100. In order to effectively train the neural ODE model, we also take the states of the ego and other vehicles as input to the neural ODE in addition to the Lidar information.

\textbf{Training data generation.} The training data comes from an integrated simulation environment (not released yet), and it is given as a ``pickle'' file. There are 200 randomly sampled trajectories and the corresponding safe controls coming from a CBF controller, and each trajectory is with 100 time-sequence of data with a discretization time of 0.1s. The Lidar information is given as a sequence of data with size 1x100, and each data point denotes a distance metric with respect to an obstacle from the angle 0 to $2\pi$. The Lidar sensing range is 20m.

During training, we normalize the Lidar information by multiplying the data with a factor of 1/200. The ego vehicle speed is also normalized by multiplying the speed with a factor of 1/180 when it is taken as an external input. The normalization of the external input is to ensure that the neural ODE can converge during training.

\textbf{Model structure.} The neural ODE (\ref{eqn:NN}) is a five-layer fully connected network with sizes 2, 64, 256, 512, and 206, respectively. The activation functions used in the model are GELU. Since we enforce the invariance on the external lidar input, we reformulate the neural ODE (\ref{eqn:NN}) into:
 \begin{equation}
    \dot {\bm x} = f_{\theta}(\bm x) + g_{\theta}(\bm x)\textbf{I},
\end{equation}

When employing feature extractors for the invariance, we use a Convolutional Neural Network (CNN) whose shape is given as $[[1, 4, 5, 2, 1], [4, 8, 3, 2, 1], [8, 12, 3, 2, 0]]$, where there are three layers, and the parameters of each layer denote input channels, output channels, kernel size, stride, and padding, respectively. After the CNN, we use a max pooling in each output channel to reduce the feature size from 100 to 12.

\textbf{Training.} The training epoch is 200 (each epoch includes the sampling of each of the 200 trajectories), and the training batch size is 20 with a batch sequence time of 10. We use RMSprop optimizer with learning rate $1e^{-3}$. The training time is about 24 hours on an RTX3090 GPU.

\begin{figure}
	\centering
	\includegraphics[scale = 0.45]{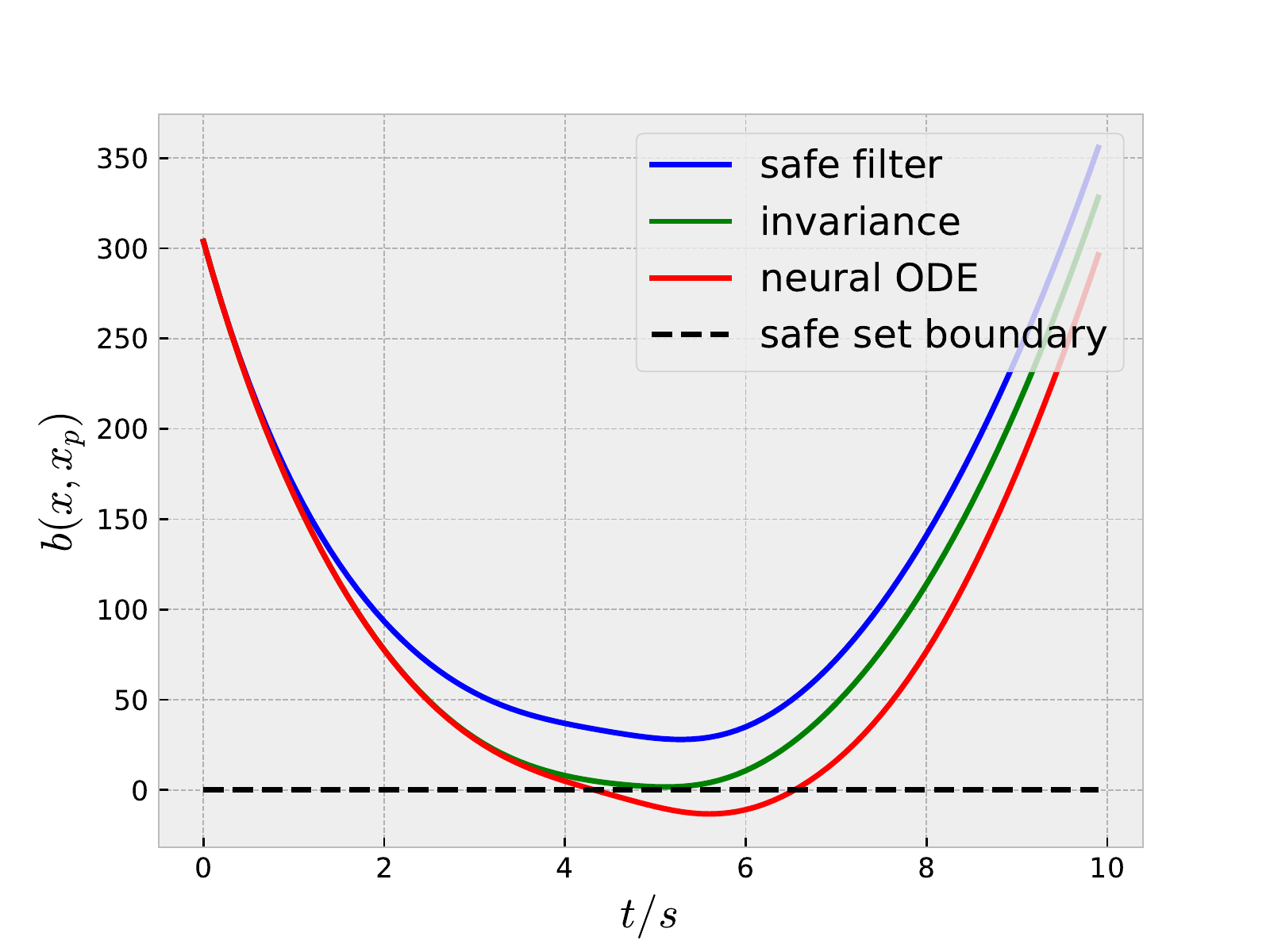} 
	\vspace{-2mm}\caption{Comparison between safe filters,  neural ODE, and invariance under a noisy Lidar point cloud. $b(\bm x, \bm x_p)\geq 0$ implies collision-free.  }%
	\label{fig:safe}%
\end{figure}

\begin{figure}
	\centering
	\includegraphics[width=0.9\columnwidth]{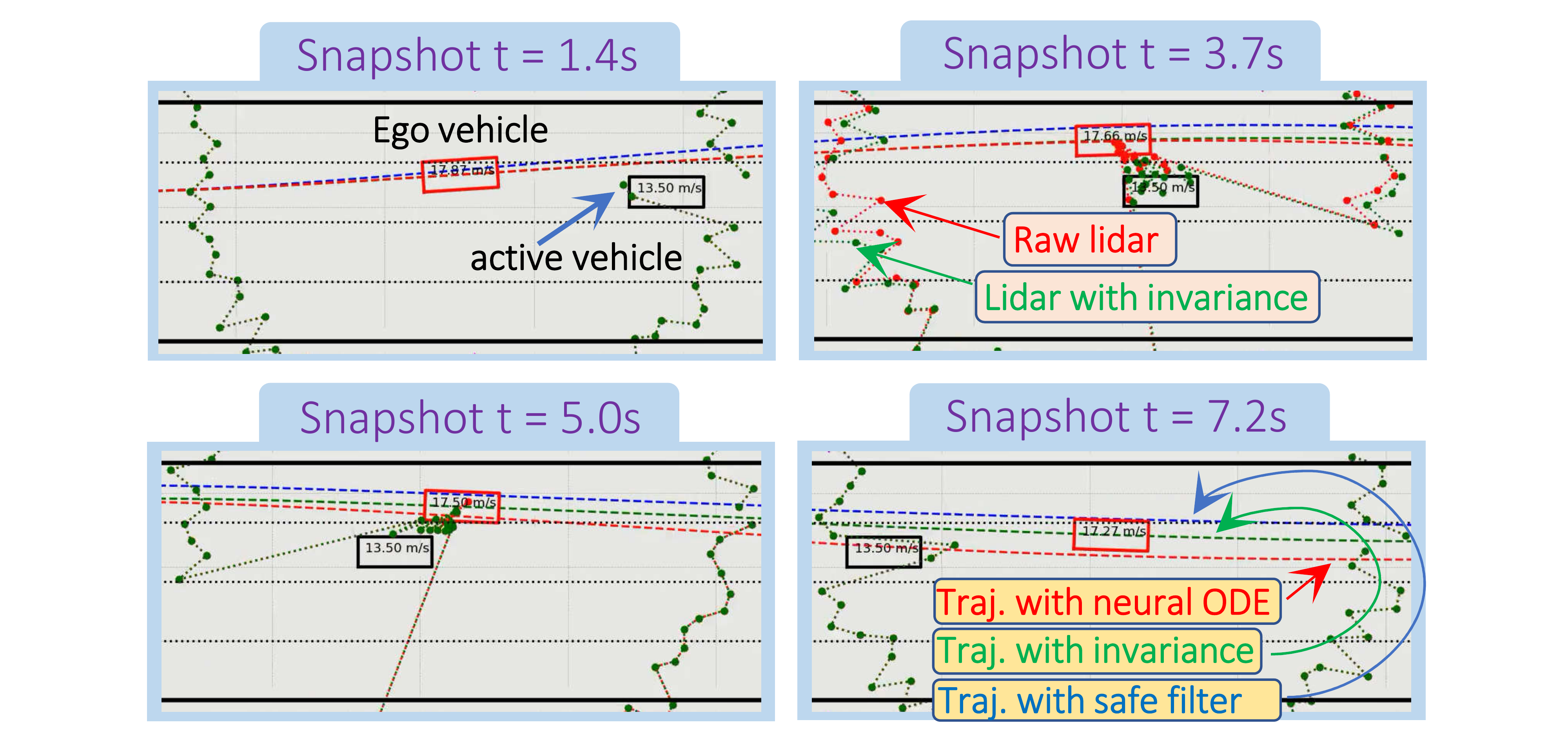} 
	\vspace{-2mm}\caption{Snapshots of simulations with trajectory comparison between safe filters,  neural ODE, and invariance under a noisy Lidar point cloud. $b(\bm x, \bm x_p)\geq 0$ implies collision-free.  A safe filter may make the ego conservative, and thus the ego stays unnecessarily far away from the ground truth, while the pure neural ODE controller may cause collision under noise.}%
	\label{fig:snap}%
\end{figure}

\textbf{Invariance v.s. safe filters v.s. pure neural ODE.}
The ego vehicle starts at a speed of $18m/s$, while the other vehicle moves at a constant speed of $13.5m/s$. In the case of noise-free Lidar sensing,  the ego may avoid collision when it overtakes the other moving vehicle with the neural ODE controller.
However, with noisy Lidar, the neural ODE controller may cause the ego vehicle to collide with the other moving vehicle during the overtaking process, as the red trajectory shown in Figure~\ref{fig:snap}. The safety constraint $b(\bm x, \bm x_p)$ becomes negative (as the red curve shown in Fig. \ref{fig:safe}) when the ego approaches the other moving vehicle, which implies collision. 

Using the proposed invariance, We map the safety requirement of the ego vehicle onto a constraint on the noisy Lidar input $\textbf{I}$. The dimension of the Lidar information is $1\times 100$, and thus, the  dimension of the decision variable of the QP that enforces the invariance is also 100. Even so, the QP can still be efficiently solved as it is just a convex optimization. With the proposed invariance, we can slightly modify the noisy Lidar data such that the outputs (controls) can guarantee the safety of the ego vehicle, as the green trajectory shown in Figure~\ref{fig:snap}. The modified Lidar information (through invariance) is illustrated by the green-dotted curve in the snapshot $t = 3.7s$ of Figure~\ref{fig:snap}. 

\textbf{Invariance with feature extractors.} In cases where the inputs of the neural ODE have high dimensions, like Lidar-based control, we may use some neural networks (such as CNN) to reduce the dimension of input features, thus reducing the complexity of the QP that enforces the invariance. For the driving example, we used a CNN to reduce the 100-dimension Lidar information to 12-dimension features, and the results are similar to the case of raw Lidar. In other words, a collision may occur when with noisy Lidar input but can be guaranteed to avoid using the proposed invariance.

\end{document}